\documentclass[lettersize,journal]{IEEEtran}
\usepackage{amsmath,amsfonts}
\usepackage{algorithmic}
\usepackage{algorithm}
\usepackage{array}
\usepackage[caption=false,font=normalsize,labelfont=sf,textfont=sf]{subfig}
\usepackage{textcomp}
\usepackage{multirow}
\usepackage{stfloats}
\usepackage{url}
\usepackage{verbatim}
\usepackage{graphicx}
\usepackage{cite}
\usepackage{colortbl}
\usepackage{color}
\usepackage{xcolor}
\usepackage{makecell}
\usepackage{booktabs}
\usepackage[breaklinks,colorlinks]{hyperref}
\usepackage{soul}
\newcommand{\PreserveBackslash}[1]{\let\temp=\\#1\let\\=\temp}

\newcolumntype{C}[1]{>{\PreserveBackslash\centering}p{#1}}
\newcolumntype{R}[1]{>{\PreserveBackslash\raggedleft}p{#1}}
\newcolumntype{L}[1]{>{\PreserveBackslash\raggedright}p{#1}}

\begin{document}
% at, but, by, for, in, nor, of, on, or, the, to , etcand up, which are usually
\title{Underwater Image Enhancement by Diffusion Model with Customized CLIP-Classifier}

\author{Shuaixin~Liu,
        Kunqian~Li*,~\IEEEmembership{Member,~IEEE},
        Yilin~Ding,
        and Qi~Qi
        % <-this % stops a space
\thanks{The research has been supported by the National Natural Science Foundation of China under Grant 62371431 and 61906177, in part by the Marine Industry Key Technology Research and Industrialization Demonstration Project of Qingdao under Grant 23-1-3-hygg-20-hy, and in part by the Fundamental Research Funds for the Central Universities under Grants 202262004.}% <-this % stops a space

\thanks{Shuaixin Liu, Kunqian Li, and Yilin Ding are with the College of Engineering, Ocean University of China, Qingdao 266404, China (liushuaixin@stu.ouc.edu.cn; likunqian@ouc.edu.cn; dingyilin@stu.ouc.edu.cn).}% <-this % stops a space

\thanks{Qi Qi is with the School of Information and Control Engineering, Qingdao University of Technology, Qingdao 266520, China (qiqi@qut.edu.cn).}% <-this % stops a space

\thanks{$^{*}$ Corresponding author: Kunqian Li (likunqian@ouc.edu.cn)}% <-this % stops a space
}

\markboth{}%
{Shell \MakeLowercase{\textit{et al.}}: Bare Dem of IEEEtran.cls for Journals}



\maketitle
\begin{abstract}
Underwater Image Enhancement (UIE) aims to improve the visual quality from a low-quality input. Unlike other image enhancement tasks, underwater images suffer from the unavailability of real reference images. Although existing works exploit synthetic images and manually select well-enhanced images as reference images to train enhancement networks, their upper performance bound is limited by the reference domain. To address this challenge, we propose CLIP-UIE, a novel framework that leverages the potential of Contrastive Language-Image Pretraining (CLIP) for the UIE task. Specifically, we propose employing color transfer to yield synthetic images by degrading in-air natural images into corresponding underwater images, guided by the real underwater domain. This approach enables the diffusion model to capture the prior knowledge of mapping transitions from the underwater degradation domain to the real in-air natural domain. Still, fine-tuning the diffusion model for specific downstream tasks is inevitable and  may result in the loss of this prior knowledge. To migrate this drawback, we combine the prior knowledge of the in-air natural domain with CLIP to train a CLIP-Classifier. Subsequently, we integrate this CLIP-Classifier with UIE benchmark datasets to jointly fine-tune the diffusion model, guiding the enhancement results towards the in-air natural domain. Additionally, for image enhancement tasks, we observe that both the image-to-image diffusion model and CLIP-Classifier primarily focus on the high-frequency region during fine-tuning. Therefore, we propose a new fine-tuning strategy that specifically targets the high-frequency region, which can be up to 10 times faster than traditional strategies. Extensive experiments demonstrate that our method exhibits a more natural appearance. The source code and pre-trained models are available on the project homepage: \url{https://oucvisiongroup.github.io/CLIP-UIE.html/}.  
\end{abstract}


% \fi
% \IEEEpeerreviewmaketitle
\section{Introduction}
% For example, Fu et al. utilize the retinex-based approach to capture a clean underwater image by combining enhanced reflectance and illumination~\cite{fu2014retinex}. Further, Zhuang et al. introduce Bayesian to improve the retinex-based algorithm estimating underwater medium transmissions~\cite{zhuang2021bayesian}.
\IEEEPARstart{W}{ith} the rise of marine ecological research~\cite{zhao2021composited,kimball2018artemis} and underwater exploration~\cite{zhou2023underwater}, there is increasing attention on processing and understanding underwater images. However, due to the unbalanced attenuation of underwater light~\cite{li2019underwater}, underwater images suffer from severe visual degradation, which often exhibits color casts, low contrast, and poor visibility~\cite{li2021underwater,jiang2023perception,qi2021underwater}. There is an urgent need for UIE techniques to improve the visual quality of underwater images. \par
Reviewing typical UIE strategies, early approaches based on physical models concentrate on accurately estimating medium transmission. This parameter represents the percentage of scene radiance reaching the camera~\cite{fu2014retinex,drews2013transmission,zhuang2021bayesian}. 
Though physical model-based methods can achieve promising performance in some cases, they tend to produce unnatural and poor results in challenging underwater scenarios~\cite{song2018rapid}. Recent approaches based on deep learning technology have gradually outperformed traditional physical model-based approaches~\cite{qi2022sguie,wang2021uiec,li2022beyond}. Although simple regression-based convolutional networks may effectively enhance images in simple underwater scenarios, they often lack generalization and robustness when dealing with complex scenes. This limitation arises from their inability to fully model the degradation specific to real-world underwater conditions. Recently, image-to-image diffusion models have demonstrated remarkable success in effectively capturing intricate empirical distributions of images. These models facilitate the seamless transformation of images across diverse domains through iterative refinement steps, thereby presenting a new approach to address such challenging problems.\par

However, image-to-image diffusion models heavily rely on paired datasets, and the inability to capture clear real-world reference images poses a challenge for solving the UIE problem~\cite{rao2023deep,shen2023udaformer,zhang2023underwater}. To tackle this issue, researchers employed various approaches to synthesize multiple enhanced images from a single image and subsequently manually select the visually optimal one as a reference~\cite{li2019underwater,qi2022sguie}. Nevertheless, the synthesized paired datasets contain subjective preferences, which are manifested in the enhanced results, leading to the natural domain shift and weakened generalization. Regarding the unsupervised or semi-supervised strategies, they have also demonstrated the capability to generate visually captivating underwater images~\cite{li2017watergan,islam2020fast}, without necessitating paired datasets or relying solely on limited paired samples; however, their performance falls short when compared to that of the fully supervised method. Another alternative approach is to combine a pre-trained diffusion model, which captures the prior knowledge of mapping transitions between different image domains, with UIE benchmark datasets, thus reducing dependency on paired datasets~\cite{du2023uiedp}. However, fine-tuning or directly continuing training with the additional UIE benchmark datasets may potentially compromise valuable prior knowledge~\cite{zhang2023adding}. Furthermore, the lack of real image guidance in the reference domain limits the quality of enhancement results, leading to unnatural and unrealistic enhanced images.\par

To address the above limitations, we propose CLIP-UIE, a new approach to treat UIE as a conditional underwater image generation problem. This approach is inspired by recent works on Denoising Diffusion Probabilistic Models~\cite{ho2020denoising,song2020denoising,dhariwal2021diffusion,saharia2022image} and Contrastive Language-Image Pretraining (CLIP) guidance for image manipulation~\cite{kim2022diffusionclip,liang2023iterative,zhou2022learning}. The proposed method is trained with a ``denoising'' objective to iteratively refine the enhancement result of the given input degraded underwater image. Initially, the key is to obtain a large number of underwater paired images. We utilize the image synthesis strategy based on color transfer~\cite{reinhard2001color,li2023tctl} to degrade in-air natural images into the real underwater domain, synthesizing the corresponding degraded underwater images, which is free of any subjective preferences. Subsequently, we train the basic image-to-image diffusion model on this large-scale synthetic paired dataset to learn the prior knowledge of mapping transitions from the underwater degradation domain to real in-air natural domain. This prior knowledge can counteract the unrealistic and unnatural synthetics effectively. It should be noted that the fine-tuning of the model to cater to downstream tasks is both inevitable and essential; however, this process may inadvertently erase the prior knowledge, resulting in overfitting and mode collapse~\cite{zhang2023adding}. Song et al.~\cite{song2020score} and Jonathan et al.~\cite{ho2022classifier} have demonstrated that a pre-trained diffusion model can be effectively conditioned by leveraging the gradients of the classifier. Thus, we can exploit a trained classifier with the prior knowledge of in-air natural domain, combined it with UIE benchmark datasets to jointly fine-tune the diffusion model to avoid overfitting and mode collapse, bridging the domain gap between synthetic and real images. \par

\begin{figure}[t]
\centering
	\includegraphics[width=1\linewidth]{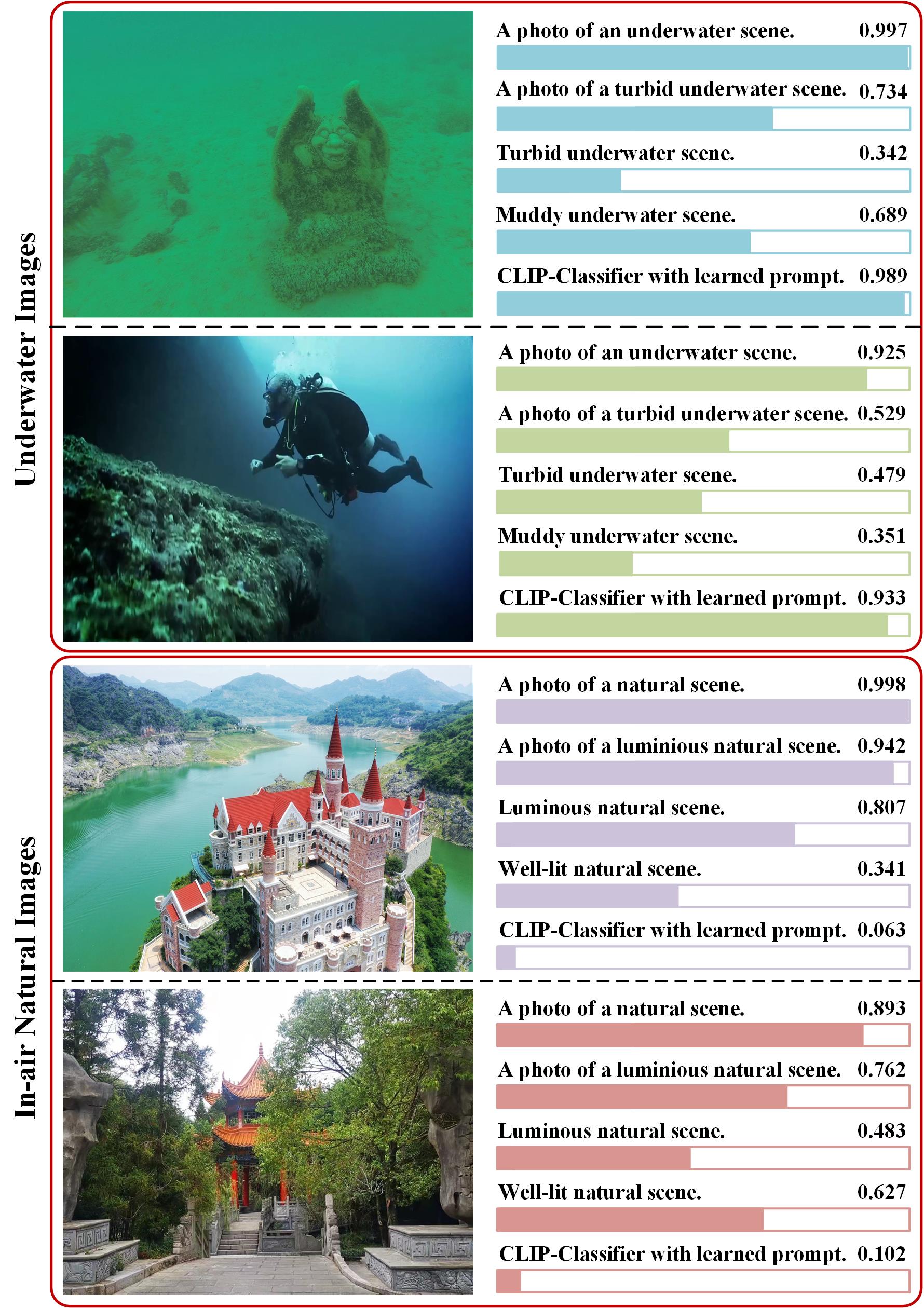}
    \caption{CLIP with proper prompts can serve as the classifier to evaluate the image quality and distinguish between in-air natural and underwater images. However, as a prevailing phenomenon, a simple approximate rephrasing of the prompt results in a significant change in CLIP score. On the contrary, our CLIP-Classifier with learned prompt (the last row of each example) shows more robust results for different types of in-air natural/underwater scenes.}\label{Fig:clip_2example}
\vspace{-1em}
\end{figure}

Recently, it has been demonstrated that Vision-Language models encapsulate generic information, that can be paired with generative model to provide a simple and intuitive text-driven interface for image generation and manipulation~\cite{gal2022stylegan}, particularly the CLIP. The utilization of CLIP as a classifier to distinguish between in-air natural and underwater degraded images is somewhat effective; however, employing its gradients directly for controlling the reverse diffusion process remains challenging. Furthermore, locating precise prompts to accurately depict various forms of underwater degradation poses a significant challenge. Additionally, as discussed in \cite{liang2023iterative}, rephrasing similar prompts often results in substantial disparities in CLIP score. Similar issues for UIE problem are illustrated in Fig. \ref{Fig:clip_2example}. We train a CLIP-Classifier that contains the prior knowledge of the in-air natural domain by constraining text-image similarity in the CLIP embedding space. Moreover, for image enhancement tasks, we find that the image-to-image diffusion model and the CLIP-Classifier mainly act in the high-frequency region during the fine-tuning process. Therefore, we propose a fast fine-tuning strategy that focuses on the high-frequency range and can be up to 10 times faster. The contributions of this paper are summarized as follows:
\begin{itemize}
     \item We propose employing color transfer techniques to simulate the degradation of $500$k in-air natural images of INaturalist~\cite{van2018inaturalist} into underwater images, guided by an authentic underwater domain, thereby addressing the limited availability of extensive underwater image samples for enhancement tasks. The image-to-image diffusion model is then trained from scratch on this large-scale synthetic dataset to obtain pre-trained model.
     \item We propose a CLIP-Classifier that inherits the prior knowledge of the in-air natural domain, and then combine it with UIE benchmark datasets to jointly fine-tune the diffusion model to mitigate catastrophic forgetting and mode collapse. Experiments and ablation studies show that the proposed CLIP-UIE performs well and breaks the limitations of the reference domain to a certain extent.
     \item We find that for image enhancement tasks, which require consistent content, the image-to-image diffusion model and the CLIP-Classifier mainly act in high-frequency regions. Therefore, we propose a new fine-tuning strategy that only focuses on high-frequency regions, significantly improving the speed of fine-tuning, even up to 10 times.
\end{itemize}

\section{Related Works}\label{Related Works}
\subsection{Underwater Image Enhancement}
In recent years, numerous approaches have been proposed in the field of underwater image enhancement, which can be classified into two distinct categories: traditional methods and deep learning-based techniques. For the traditional methods, researchers focus on directly adjusting image pixel values to produce a subjectively and visually appealing image, using techniques such as image fusion~\cite{zhang2024underwater}, histogram equalization~\cite{zhang2021enhancing}, and pixel distribution~\cite{zhou2024pixel}, etc. Ancuti et al.~\cite{ancuti2017color} propose the fusion-based underwater image enhancement method, exploiting two images derived from a color-compensated and white-balanced version to prompt color contrast. However, these traditional methods are not robust to different types of underwater degradation~\cite{kimball2018artemis}, due to the omission of the underwater imaging mechanism. To improve the visual quality, physical model-based methods directly estimate the parameters of underwater imaging models based on prior information. For example, Fu et al.~\cite{fu2014retinex} utilize the retinex-based approach to capture a clean underwater image by combining enhanced reflectance and illumination. To further promote the performance, Zhuang et al.~\cite{zhuang2021bayesian} improve the retinex-based algorithm by introducing Bayesian to estimate underwater medium transmissions. Nevertheless, owing to the scarcity of prior information, the enhanced image easily becomes unnatural and over/under-enhanced.\par

Recently, deep learning has made significant strides in the fields of computer vision and image processing, leading to a series of endeavors aimed at enhancing the performance of underwater image enhancement through data-driven strategies. However, underwater images suffer from the unavailability of real reference images. Concretely, Li et al.~\cite{li2019underwater} construct an underwater image benchmark (UIEB) by selecting the best result of various approaches as the reference. However, the performance of existing UIE methods, especially deep learning-based data-driven approaches, is limited by the unavoidable bias of those synthetic alternatives. Several GAN-based methods \cite{li2017watergan,islam2020fast} have been developed to generate pleasing underwater images, relaxing the requirement of paired underwater images. However, the performance of these methods lacks stability and generally falls behind fully supervised approaches, often introducing unrealistic elements and visual distortions.\par

Utilizing in-air natural images to synthesize paired underwater degradation datasets provides a promising solution. Du et al.~\cite{du2023uiedp} transform in-air natural images into the pseudo underwater image domain by Koschmieder's light scattering model, utilizing the prior knowledge of in-air natural images to counteract the adverse impact of unreal synthetic. However, the pseudo underwater domain based on the light scattering model sometimes cannot model the complexity of real-world underwater degradation. As a more intuitive and efficient alternative, we propose a new synthesis strategy based on color transfer~\cite{reinhard2001color} to generate paired datasets by degrading natural in-air images to the real underwater domain. This approach introduces additional prior knowledge: the mapping transitions between the real underwater degradation domain and the real in-air natural domain.

% The work by Tang et al.~\cite{tang2023underwater} applies diffusion models to enhance underwater images conditioned on degraded underwater inputs, directly using the UIEB benchmark. More closely related to ours is the work of Du et al.~\cite{du2023uiedp}, where the diffusion model is improved by capturing natural image priors, counteracting the adverse impact of unreal synthetic. In particular, they transform the natural images into the pseudo underwater image domain by estimating the medium transmission map using a light scattering model.

\subsection{Diffusion Models}
In recent studies, diffusion models have shown impressive performance in image generation by inverting image noising process using reverse-time stochastic differential equations (SDEs)~\cite{gal2022image}. However, this inversion process tends to alter the image context. To condition the denoising process, Classifier Guidance~\cite{ho2022classifier,song2020score} is an effective method to improve the sample quality of conditional diffusion models by utilizing the gradients of the classifier. An alternative strategy is to establish the relationship between input and conditional information in the latent space of the model~\cite{avrahami2022blended,saharia2022image}. Controlling diffusion models facilitates personalization, customization, or task-specific image generation. Hence, the diffusion models have been successfully applied to various image enhancement tasks. The image-to-image diffusion model was first introduced by Saharia et al.~\cite{saharia2022image} and has been exploited for image enhancement~\cite{tang2023underwater,du2023uiedp}, inpainting~\cite{saharia2022palette}, and super-resolution~\cite{saharia2022image}, etc.\par
Conditional diffusion models have the ability to acquire mapping transitions between diverse image domains~\cite{zhang2023adding}. Building upon this observation, we train the conditional diffusion model on the synthetic dataset produced by color transfer to capture the prior knowledge of mapping transitions between the real underwater degradation domain and the real in-air natural domain. And then we will use a classifier to retain this valuable prior knowledge in subsequent fine-tuning of the pre-trained diffusion model, bridging the domain gap between synthetic and real images.  
% Recent studies show that wherein a pre-trained diffusion model can be conditioned using the gradients of a classifier.
% Conditional diffusion models can learn the mapping between different image domains~\cite{zhang2023adding}. Based on this observation, Du et al.~\cite{du2023uiedp} transform natural images into the pseudo underwater image domain by Koschmieder’s light scattering model, and then make full use of the natural image priors to train the diffusion model. However, the degradation domain based on the light scattering model cannot fully model the complexity of real-world underwater degradation. In contrast, our proposal uses the color transfer to synthetic images mentioned above, and the conditional diffusion model is trained to learn the mapping between the real underwater degradation domain and the real natural domain. 

\subsection{CLIP-based Image Manipulation}
The CLIP model comprises a text encoder and an image encoder, which are jointly trained \cite{radford2021learning}. Leveraging the knowledge from 400 million carefully curated image-text pairs, it demonstrates the capability to accurately predict pairings of unseen \texttt{(image, text)} examples in a given batch. It has shown remarkable performance in text-to-image generation. Gal et al.~\cite{gal2022stylegan} and Patashnik et al.~\cite{patashnik2021styleclip} use CLIP models to modify StyleGAN-generated images with text prompts. More recently, Kim et al.~\cite{kim2022diffusionclip} and Avrahami et al.~\cite{avrahami2022blended} propose methodologies for leveraging the diffusion model guided by CLIP models in order to achieve global text-to-image synthesis, as well as facilitate local image editing. Text-driven generative models demonstrate strong performance when provided with explicit word prompts, such as ``apple'', ``banana'', and ``orange'', which pertain to objective entities. However, it remains challenging to identify precise prompts for abstract image styles and visual qualities. Moreover, prompt engineering for abstract concepts necessitates the involvement of domain experts to meticulously annotate each image, and a notable challenge lies in the fact that similar prompts may yield disparate CLIP scores~\cite{zhou2022learning} (see Fig.~\ref{Fig:clip_2example}). It restricts the user to generating and manipulating custom domains for abstract semantic concepts using the pre-trained CLIP.

To address the aforementioned issues, recent emerging research suggests fine-tuning the CLIP model with ample training data for domain adaptation \cite{gal2022stylegan}; however, this approach is labor-intensive and time-consuming. Concurrently, prompt learning with frozen CLIP model is the alternation to extract accurate low-level image representations~\cite{zhou2022learning}, such as luminance, exposure, and contrast~\cite{liang2023iterative}. The present study also employs prompt learning to train a CLIP-Classifier, enabling the discrimination between in-air natural and underwater degraded images. Subsequently, the gradients derived from this CLIP-Classifier are employed to guide the sampling process of the diffusion model.

\section{Preliminary Knowledge and Proposed Method}
\subsection{Preliminary Knowledge of Diffusion Models}
Diffusion probabilistic models are a type of latent variable models that define a diffusion Markov chain to slowly introduce random noise into the data, and a corresponding reverse process to gradually remove the noise~\cite{ho2020denoising}. The diffusion process is a forward process that gradually injects noise into the data sample ${x}_{0}$ over $T$ time steps, diffusing the data samples through Gaussian transition:
\begin{equation}
  q(x_{t}|x_{t-1})=\mathcal{N}(x_{t};\sqrt{1-\beta_{t}}x_{t-1}, \beta_{t}\mathbf{I}),
\end{equation}
where $\left \{ \beta _{t}  \right \}_{t=1}^{T}$ is an increasing variance schedule with values of $(0,1)$. The intermediate variable ${x}_{t}$ can be expressed as:
\begin{equation}
\label{eq:x_t_produce}
  x_{t}=\sqrt{\bar{\alpha}_{t}}x_{0}+\sqrt{1-\bar{\alpha}_{t}}\epsilon,  \epsilon \sim \mathcal{N}(0,\mathbf{I}),
\end{equation}
where $\alpha _{t} =1-\beta _{t} $, $\bar{ \alpha }_{t}= {\textstyle \prod_{i=1}^{t} \alpha _{i}}$. The final diffuse distribution sample ${x}_{T}$ is prior random Gaussian noise.

To recover the data from the random noise ${x}_{T}$ mentioned above, the reverse process involves iterative denoising~\cite{ho2020denoising}, which is parametrized by another Gaussian transition:
\begin{equation}
  p_{\theta } (x_{t-1}|x_{t})=\mathcal{N}(x_{t-1};\mu _{\theta }(x_{t},t), \sigma _{\theta }(x_{t}, t)\mathbf{I}) ,
\end{equation}
where $\mu _{\theta }(x_{t},t)$ and $\sigma _{\theta }(x_{t}, t)$ are the functions of mean and variance, respectively. Some works~\cite{ho2020denoising,luo2022understanding} set the two Gaussian variances of the forward and reverse processes to match exactly, and the diffusion model just need to optimize $\mu _{\theta }(x_{t},t)$. Then, when $x_{t}$ is available, $\mu _{\theta }(x_{t},t)$ can be reparameterized as: 
\begin{equation}
  \mu _{\theta }(x_{t},t)=\frac{1}{\sqrt{\alpha _{t}}}(x_{t}-\frac{\beta _{t}}{\sqrt{1-\bar{\alpha}_{t}} }\epsilon_{\theta}(x_{t},t)), 
\end{equation} 
where $\epsilon_{\theta}(x_{t},t)$ is a neural network similar to a function approximator trained to predict $\epsilon$ from $x_{t}$, and the optimizing objective of Evidence Lower Bound (ELBO) is converted from $\mu _{\theta }(x_{t},t)$ to $\epsilon_{\theta}(x_{t},t)$, as follows:   
\begin{equation}
\label{eq:ELBO}
  \mathcal{L}_{1}=\min_{\theta } \mathbb{E}_{x_{0}\sim q(x_{0}),\epsilon \sim \mathcal{N}(0,\mathbf{I}),t}\left \|\epsilon-\epsilon_{\theta}(x_{t},t)\right \|^{2}. 
\end{equation}

As another understanding of the diffusion model, Song et al.~\cite{song2020score,song2019generative} unify the diffusion model, a discrete multi-step denoising process, into a special form of a continuous stochastic differential equation (SDE), and the reverse process is to simulate the score of each marginal distribution $\bigtriangledown\log p(x_{t})$ from $T$ to $0$. Recall that $\bigtriangledown$ is shorthand for $\bigtriangledown_{x_{t}}$ in the interest of brevity. Meanwhile, the score matching of the score-based generative model is equivalent to ELBO optimization, and the score $\bigtriangledown\log p(x_{t})$ can be represented as the linear mapping of $\epsilon_{\theta}(x_{t},t)$ with the following relationship:
\begin{equation}
\label{eq:linear_mapping}
  \bigtriangledown\log p(x_{t})=-\frac{1}{\sqrt{1-\bar{\alpha}_{t} }}\epsilon_{\theta}(x_{t},t).
\end{equation}

\begin{figure*}[t]
\centering
	\includegraphics[width=0.98\linewidth]{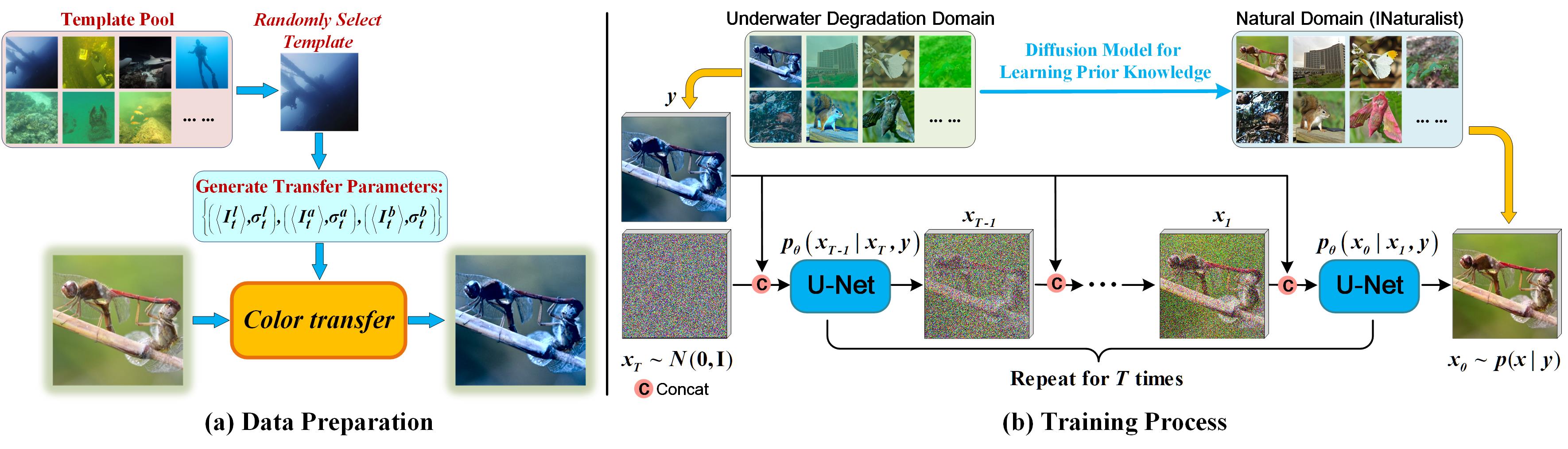}
    \caption{The preparation for the pre-trained model. (a) Randomly select a \textbf{Template} from the \textbf{Template Pool} (underwater domain). Then, the \textbf{Color Transfer} module, guided by the template, degrades an in-air natural image from INaturalist~\cite{van2018inaturalist} into underwater domain, constructing paired datasets for training image-to-image diffusion model. (b) The image-to-image diffusion model \textbf{SR3}~\cite{saharia2022image} is trained to learn the prior knowledge of mapping transitions from the real underwater degradation domain to the real in-air natural domain and to generate the corresponding enhancement results based on input synthetic underwater images produced by Color Transfer.}
    \label{Fig:pipeline}
\end{figure*}

For conditional generation models, Dhariwal et al.~\cite{dhariwal2021diffusion} uses gradients from a classifier $p(y|x_{t})$ to guide the reverse process, where $y$ is the condition information. By Bayes formula, the score function of this is formed as 
\begin{equation}
  \label{eq:score_condition}
  \begin{aligned}
    \bigtriangledown\log p(x_{t}|y)&=\bigtriangledown \log (\frac{p(x_{t})p(y|x_{t})}{p(y)})\\ 
    =\bigtriangledown&\log p(x_{t})+\bigtriangledown\log p(y|x_{t})-\bigtriangledown\log p(y)\\
    =\bigtriangledown&\log p(x_{t})+\bigtriangledown\log p(y|x_{t}). 
  \end{aligned}
\end{equation}

Notably, the gradient of $\bigtriangledown\log p(y)$ with respect to $x_{t}$ is zero, $\bigtriangledown\log p(x_{t})$ is defined as an unconditional score, and $\bigtriangledown\log p(y|x_{t})$ is defined as the adversarial gradient~\cite{luo2022understanding}. Then, $\epsilon_{\theta}(x_{t},t)$ is redefined as $\epsilon_{\theta}(x_{t},y,t)$ which corresponds to the score of the joint distribution~\cite{dhariwal2021diffusion}:
\begin{equation}
  \epsilon_{\theta}(x_{t},y,t)=\epsilon_{\theta}(x_{t},t)-\sqrt{1-\bar{\alpha}_{t}}\bigtriangledown\log p(y|x_{t}).
\end{equation}

Simultaneously, the optimizing objective is rewritten as: 
\begin{equation}
\label{eq:leaning_objective_y}
\mathcal{L}_{2}=\min_{\theta } \mathbb{E}_{x_{0}\sim q(x_{0}),\epsilon \sim \mathcal{N}(0,\mathbf{I}),y,t}\left \|\epsilon-\epsilon_{\theta}(x_{t},y,t)\right \|^{2}.
\end{equation}

After training, the sampling procedure for regular DDPM~\cite{ho2020denoising} or DDIM~\cite{song2020denoising} can be used. It is found that Eq. (\ref{eq:score_condition}) provides a theoretical basis for multi-guidance when conditions are independent of each other.

\subsection{Pipeline of Underwater Image Enhancement}
\subsubsection{\textbf{Multi-Guidance for Diffusion Models}}
\label{se:multi-classifiers}
Our method leverages a diffusion model for underwater image enhancement, and we use the image-to-image model SR3~\cite{saharia2022image} as the base diffusion model framework. To condition the SR3 on the input underwater image \textbf{$y$}, the input \textbf{$y$} is concatenated with the intermediate variable \textbf{$x_{t}$} along the channel dimension (see in Fig.~\ref{Fig:pipeline} (b)), and experiments have shown that simple concatenation produces similar generation quality and enhances the model's robustness~\cite{saharia2022image}. \par

The most prevalent form of guidance for conditional diffusion models is Classifier Guidance \cite{ho2022classifier,song2020score}, which is derived from score-based generation models. Our goal is to train a neural network to predict $\bigtriangledown\log p(x_{t}|y)$, the score of the conditional model, at arbitrary transition steps. Recall the score function from Eq. (\ref{eq:score_condition}):
\begin{equation}
    \bigtriangledown\log p(x_{t}|y)
    =\bigtriangledown\log p(x_{t})+\bigtriangledown\log p(y|x_{t}),
    \nonumber
\end{equation}
where $y$ represents the conditional information, which is an underwater image in this work, and of course, it can be other conditions like prompts, Canny edges, human poses, etc~\cite{zhang2023adding}. Additionally, a classifier $p(y|x_{t})$ is used to take in arbitrary noisy $x_{t}$ and attempt to predict conditional information $y$. Directly using the gradients $\bigtriangledown\log p(x_{t}|y)$ constructed from the underwater image $y$ and the manually selected reference image $x_{0}$ to control SR3 may produce competitive enhancement results, but its upper performance bound is limited by the reference domain. The common approach to address this problem is to leverage the prior knowledge from pre-trained models to counteract the limitation of the reference domain. However, directly fine-tuning diffusion models with UIE benchmark datasets tends to compromise this prior knowledge and makes it difficult to fully exploit the diffusion models' potential.\par
In Classifier Guidance~\cite{song2020score,luo2022understanding}, the score $\bigtriangledown\log p(x_{t}|y)$ can be interpreted as learning an unconditional score function $\bigtriangledown\log p(x_{t})$ combined with the adversarial gradients of a classifier $p(y|x_{t})$. Building on this observation, we can easily combine Bayes formula to expand a single classifier into multiple classifiers and achieve multi-guidance of diffusion models. This discover offers a fresh perspective, enabling us to introduce more prior knowledge of different domains in the form of classifiers to jointly fine tune the diffusion model, thereby breaking the limitation of the reference domain. Specially, we divide our approach, abbreviated as CLIP-UIE, into two stages. In the first stage, we use the synthetic dataset produced by color transfer as the single conditional information to guide the image-to-image diffusion model SR3. This approach aims to enable SR3 to capture the prior knowledge of mapping transitions from the underwater degradation domain to in-air natural domain. In the second stage, we exploit a trained classifier that possesses the prior knowledge of the in-air natural domain. Then we combine it with the UIE benchmark dataset to jointly fine-tune the pre-trained baseline SR3, guiding the enhancement results towards the in-air natural domain and bridging the domain gap between synthetic and real images.\par 

\textbf{Stage-one: Obtain the Pre-trained Model with Prior Knowledge of Domain Adaption.} 

\textit{\textbf{(a) Data Preparation:}} Natural images can be transformed into underwater images through a light scattering model~\cite{du2023uiedp}. However, the degradation type of this kind of synthetic images is relatively simple and limited in number, which cannot simulate diverse underwater scenes. Considering that color transfer can efficiently and intuitively translate between image domains by transferring the color characteristics of one image to another~\cite{reinhard2001color}, it enlightens and enables us to transform natural images captured in air into corresponding degraded underwater images. The more images of different underwater degradation types we provide, the closer the synthetic dataset gets to the real underwater domain. \par

Recall that the color transfer align the color appearance of a source image $I_{s}$ with a target image $I_{t}$ (all images have been converted from RGB to CIELAB color space ($L^{\ast}a^{\ast}b^{\ast}$)):
\begin{equation}
\label{eq:color_transfer}
\begin{aligned}
l_{s}'&=\frac{\sigma_{t}^{l}}{\sigma_{s}^{l}}(l_{s}-\left \langle  I_{s}^{l} \right \rangle )+\left \langle I_{t}^{l} \right \rangle,\\
a_{s}'&=\frac{\sigma_{t}^{a}}{\sigma_{s}^{a}}(a_{s}-\left \langle  I_{s}^{a} \right \rangle )+\left \langle I_{t}^{a} \right \rangle,\\ 
b_{s}'&=\frac{\sigma_{t}^{b}}{\sigma_{s}^{b}}(b_{s}-\left \langle  I_{s}^{b} \right \rangle )+\left \langle I_{t}^{b} \right \rangle,
\end{aligned}   
\end{equation}
where $\left \langle \cdot  \right \rangle $ refers to calculate the mean value of the channel, and $\sigma_{s}^{i}$, $\sigma_{t}^{i}$, $i\in [l,a,b]$ are the standard deviations of the different channels in the source and target images, respectively. $l_{s}$, $a_{s}$ and $b_{s}$ are the pixel values of the source image, and the output $l_{s}'$, $a_{s}'$ and $b_{s}'$ are the pixel values of the resulting image.\par

In our case, for each in-air natural image $I_{s}$, we randomly select one image from \textbf{Template Pool} (the underwater scene domain) to serve as the target image $I_{t}$, which also functions as \textbf{Template}. We then use $I_{t}$ to guide $I_{s}$ in transitioning to the underwater scene domain via Eq. (\ref{eq:color_transfer}). The detailed process is depicted in Fig.~\ref{Fig:pipeline} (a). Finally, we convert the degraded synthetic image back to RGB color space. To facilitate the distinction, our synthetic dataset created by color transfer is referred to as the \textbf{UIE-air dataset}, while the synthetic dataset obtained by manual selection is referred to as the \textbf{UIE-ref dataset}, e.g., UIE benchmark datasets \textbf{UIEB}~\cite{li2019underwater} and \textbf{SUIM-E}~\cite{qi2022sguie}. \par

\textit{\textbf{(b) Training Procedure:}} Given the paired \textbf{UIE-air dataset} denoted as $\mathcal{D}=\left \{ (x_{i}, y_{i} ) \right \}_{i=1}^{N}$, our objective is to train a conditional diffusion model that implies the prior knowledge of mapping transitions from the underwater degradation domain to the real in-air natural domain. We aim to achieve this by minimizing the learning objective $\mathcal{L}_{2}$ (refer to Eq. (\ref{eq:leaning_objective_y})) over an in-air natural image $x$, conditioned on a synthesized underwater image $y$ (as shown in Fig.~\ref{Fig:pipeline} (b)). The gradients from the classifier $p(y|x_{t})$ are used directly to guide the reverse process, ensuring the sample $x_{t}$ adhere to the conditioning information $y$. After the training is completed, the prior knowledge of the pre-trained model is utilized to mitigate the adverse impact of unreal synthetic. Subsequently, this model will be guided by additional \textbf{UIE-ref dataset} to cater to specific underwater scenarios, but this fine-tuning or directly continue training may inadvertently compromise the prior knowledge, leading to catastrophic forgetting and mode collapse.\par

To avoid this issue, thanks to the Classifier Guidance strategy, we can train a classifier to possesses the prior knowledge of the in-air natural domain, distinguishing between in-air natural domain images and underwater domain images. Then, we exploit this classifier along with the additional \textbf{UIE-ref dataset} to jointly fine-tune the diffusion model. This classifier helps retain the prior knowledge throughout the fine-tuning process.\par
\begin{figure*}[t]
\centering
	\includegraphics[width=1\linewidth]{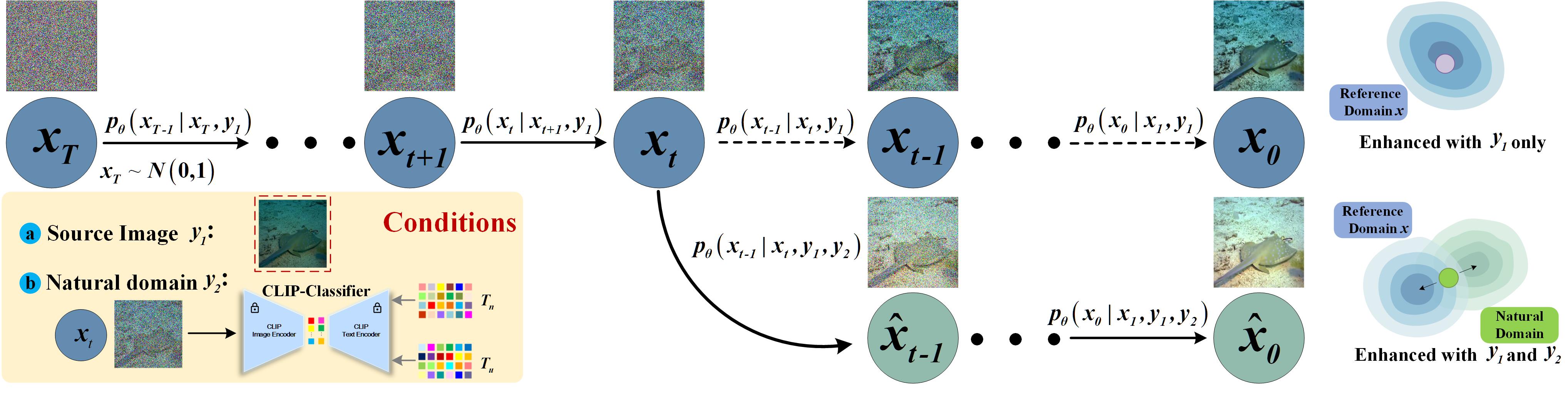}
    \caption{\textbf{New fine-tuning strategy}. From state $x_{T}$, according to Eq.~\ref{eq:leaning_objective_y}, we first adopt a single classifier guidance strategy, using condition $y_{1}$---the input source image---to guide the reverse diffusion process until state $x_{t}$. Then, we switch to the multi-classifier guidance strategy according to Eq.~\ref{eq:multi-classifiers_learning_objective}. With multi-condition guidance, the intermediate results from $x_{t}$ to $x_{0}$ move towards to the in-air natural domain, mitigating the damage of fine-tuning to the prior knowledge of the pre-trained model.}
    \label{Fig:clip_guidance}
\end{figure*}

\begin{figure}[t]
\centering
	\includegraphics[width=1\linewidth]{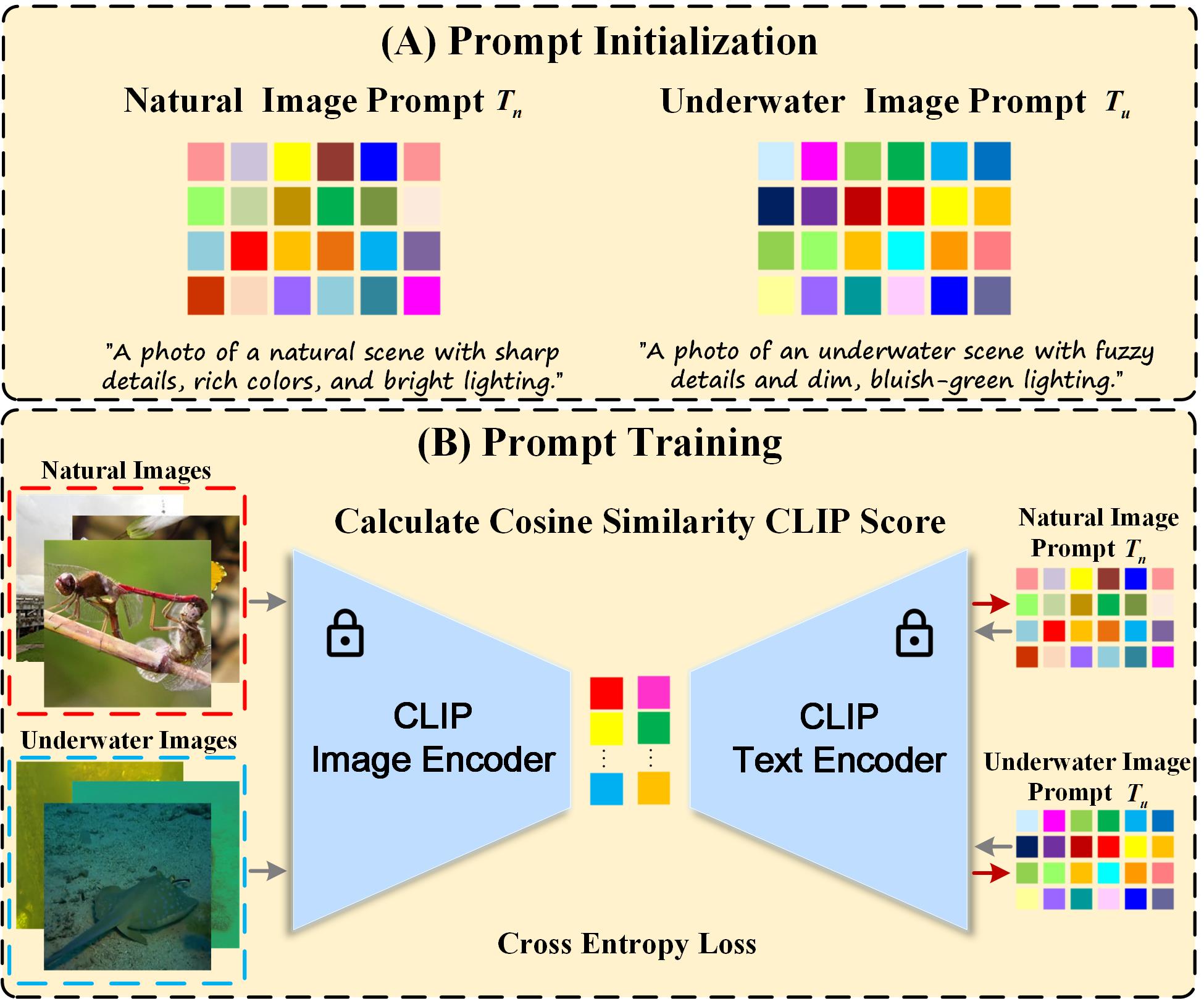}
    \caption{Illustration of the prompt learning for CLIP-Classifier. (A) \textbf{Prompt Initialization}. Given two text prompts describing the in-air natural image $I_{n}$ and underwater image $I_{u}$. We encode each text to get the initial in-air natural image prompt $T_{n}\in\mathbb{R}^{N\times512}$ and the initial underwater image prompt $T_{u}\in\mathbb{R}^{N\times512}$. (B) \textbf{Prompt Training}. We use the cross-entropy loss to constrain the learnable prompts, aligning learnable prompts and images in the CLIP latent space by maximizing the cosine similarity for matched pairs $\left \{T_{n}, I_{n}\right \}$ and $\left \{T_{u}, I_{u}\right \}$. The base model of CLIP is frozen throughout. The further use of learned prompt for CLIP-Classifier can be seen in Fig.~\ref{Fig:clip_guidance}.}
    % This also means maximizing the distance between the representation of natural domain and underwater domain in the CLIP latent space.
    \label{Fig:prompt_learning}
\end{figure}
\textbf{Stage-two: Fine-tune the Pre-trained Model under Classifier Guidance.} 

Using Bayes formula, the score function of the multi-condition model under Classifier Guidance can be derived as:
\begin{equation}
\label{eq:score_condition_multi}
\begin{aligned}
    \bigtriangledown\log p(x_{t}|y_{1},\dots,&y_{m})=\bigtriangledown \log (\frac{p(x_{t})p(y_{1},\dots ,y_{m})}{p(y_{1},\dots ,y_{m})})\\ 
    =\bigtriangledown&\log p(x_{t})+\bigtriangledown\log p(y_{1},\dots ,y_{m}|x_{t}).
  \end{aligned}
\end{equation}\par

We assume that the conditions are independent of each other, and further derive the following formula:
\begin{equation}
  \label{eq:independent}
\begin{aligned}
    \bigtriangledown\log p(y_{1},y_{2}|x_{t})=\bigtriangledown\log p(y_{1}|x_{t})+\dots +\bigtriangledown\log p(y_{m}|x_{t}).
\end{aligned}
\end{equation}\par
In our UIE task, there are only two conditions, the source image $y_{1}$ and the in-air natural domain $y_{2}$. So we simplify Eq. (\ref{eq:independent}) by setting $m=2$. Then, substituting this into Eq. (\ref{eq:score_condition_multi}), we have:
\begin{equation}
\label{eq:multi_guidance}
\begin{aligned}
    \bigtriangledown\log p(x_{t}|y_{1},y_{2})=&\bigtriangledown\log p(x_{t})+\bigtriangledown\log p(y_{1}|x_{t})\\&+\bigtriangledown\log p(y_{2}|x_{t}),
\end{aligned}
\end{equation}
where $p(y_{1}|x_{t})$ is the same classifier as $p(y|x_{t})$ in Eq. (\ref{eq:score_condition}) that makes the sample $x_{t}$ adhere to the source image $y_{1}$. The $p(y_{2}|x_{t})$ is the trained classifier that makes the sample $x_{t}$ move towards the in-air natural domain $y_{2}$.\par 

To introduce fine-grained control to either encourage or discourage the model to move in the direction close to the in-air natural domain, we use a hyperparameter term $\lambda \in [0,1]$ to scale the adversarial gradient of the classifiers as follows: 
\begin{equation}
\label{eq:multi_guidance_scale}
\begin{aligned}
    \bigtriangledown\log p(x_{t}|y_{1},y_{2})=&\bigtriangledown\log p(x_{t})+\lambda \bigtriangledown\log p(y_{1}|x_{t})\\&+(1-\lambda )\bigtriangledown\log p(y_{2}|x_{t}),
\end{aligned}
\end{equation}
when $\lambda <0.5$, the conditional diffusion model prioritizes the conditioning information $y_{2}$ and moves more towards the in-air natural domain (see in Fig.~\ref{Fig:clip_guidance}). The value of $\lambda$ is empirically set to 0.4 in our experiments. Meanwhile, according to Eq. (\ref{eq:linear_mapping}) and Eq. (\ref{eq:leaning_objective_y}), we rewrite $\epsilon_{\theta}(x_{t},t)$ as $\epsilon_{\theta}(x_{t},y_{1},y_{2},t)$, combining the adversarial gradient of the two classifiers:
\begin{equation}
\label{eq:noise_scale}
\begin{aligned}
    \epsilon_{\theta}(x_{t},y_{1},y_{2},t)=&\epsilon_{\theta}(x_{t},t)-\lambda  \sqrt{1-\bar{\alpha}_{t}}\bigtriangledown\log p(y_{1}|x_{t})\\
&-(1-\lambda  )\sqrt{1-\bar{\alpha}_{t}}\bigtriangledown\log p(y_{2}|x_{t}),
\end{aligned}
\end{equation}
and the overall learning objective $\mathcal{L}_{3}$ of the multi-guidance diffusion model is depicted in Eq. (\ref{eq:multi-classifiers_learning_objective}):
\begin{equation}
\label{eq:multi-classifiers_learning_objective}
\begin{aligned}
\mathcal{L}_{3}=\min_{\theta } \mathbb{E}_{x_{0}\sim q(x_{0}),\epsilon \sim \mathcal{N}(0,\mathbf{I}),y_{1},y_{2},t}\left \|\epsilon-\epsilon_{\theta}(x_{t},y_{1},y_{2},t)\right \|^{2}.
\end{aligned}
\end{equation}\par
This learning objective $\mathcal{L}_{3}$ is directly used in fine-tuning the pre-trained diffusion model. With multi-condition guidance, the intermediate results from $x_{t}$ to $x_{0}$ are constrained by both the source image $y_{1}$ and in-air natural domain $y_{2}$, mitigating the damage that fine-tuning can cause to the prior knowledge of the pre-trained model.

\subsubsection{\textbf{Prompt Learning for CLIP-Classifier}}
In Section \ref{se:multi-classifiers}, what we want is to construct a classifier $p(y_{2}|x_{t})$ with conditional information $y_{2}$ (in-air natural domain) and then combine it with the classifier $p(y_{1}|x_{t})$ with conditional information $y_{1}$ (\textbf{UIE-ref dataset}). This combination will jointly control the diffusion model's generation process, making the result move toward the in-air natural domain. Thanks to the CLIP's near-perfect text-driven property~\cite{radford2021learning}, we design a CLIP-Classifier to implement this vision, and all we need to do is to find prompts that can distinguish between in-air natural images and underwater images. However, prompts for the quality and abstract perception of in-air natural and underwater images often require a lot of manual prompt tuning, and the similar prompts easily lead to huge differences in CLIP score, as shown in Fig.~\ref{Fig:clip_2example}. To overcome this issue, CoOp~\cite{zhou2022learning} models a prompt's context words with learnable vectors, extending the application of  CLIP-like vision-language models to downstream tasks. For our task, we only need two prompts to characterize in-air natural and underwater images. Therefore, similar to CLIP-LIT~\cite{liang2023iterative}, we utilize the prompt learning to model a prompt's text with the longest learnable tensors, maximizing the ability to characterize images. Note that the base model of CLIP is frozen in the entire process. The process of prompt learning involves two key components: prompt initialization and prompt training, as detailed below.

\textbf{Prompt Initialization.} Randomly given two texts describing in-air natural and underwater images respectively, CLIP converts each text into the multi-model embedding space, obtaining the in-air natural image prompt $T_{n}\in\mathbb{R}^{N\times512}$, and the underwater image prompt $T_{u}\in\mathbb{R}^{N\times512}$ (refer to Fig.~\ref{Fig:prompt_learning} (a)). $N$ represents the length of prompt tokens, with an upper limit of 77. CoOp~\cite{zhou2022learning} has experimentally shown that more prompt tokens led to better performance. Thus, we model $T_{n}$ and $T_{u}$ as learnable tensors with $N=77$, maximizing the length of prompt tokens.

\textbf{Prompt Training.} Given an in-air natural image $I_{n}$ and an underwater image $I_{u}$, the image encoder $\Phi(\cdot)$ of the frozen CLIP takes them as the input and extracts image features. On the other hand, by forwarding the prompt $T_{n}$ and the underwater prompt $T_{u}$ to the text encoder $\Psi(\cdot )$, we can obtain the text features. Given a batch of image-text pairs, the prediction probability is calculated as follows: 
\begin{equation}
 P(T_{i}\mid I)=\frac{e^{cos(\Psi(T_{i}), \Phi(I))}}{\sum _{i\in\left \{ n, u \right \} } e^{cos(\Psi(T_{i}), \Phi(I))}}, 
\end{equation}
where $I$ is an image, $I\in\left \{ I_{n},I_{u}\right \} $. $cos(\cdot ,\cdot )$ denotes cosine similarity. Specially, the $T_{n}/T_{u}$ is shared among all in-air natural/underwater images. Similar to CLIP~\cite{radford2021learning}, the prompt learning is trained to align learnable prompts and images in embedding spaces by maximizing the cosine similarity for matched pairs $\left \{T_{n}, I_{n}\right \}$ and $\left \{T_{u}, I_{u}\right \} $, while minimizing the cosine similarity for incorrect pairs. The cross-entropy loss can be used as a learning objective, and its gradients can be propagated all the way back to update the prompt tensors (see in Fig.~\ref{Fig:prompt_learning} (b)). Since there are only two prompts in our prompt engineering, we can simplify the process by directly using the binary cross-entropy to classify in-air natural and underwater images, thereby learning the prompt tensors:
\begin{equation}
\mathcal{L}_{p}=-(q\ast logP(T_{n}\mid I)+(1-q)\ast log(1-P(T_{n}\mid I))), 
\end{equation}
where $q$ is the label of one-hot, which is $1$ for in-air natural images and $0$ for underwater images.

\textbf{CLIP-Classifier.} After training, we can get the well-performing prompts and then construct the CLIP-Classifier by measuring the similarity between the generation image and the prompts in the CLIP space: 
\begin{equation}
 \mathcal{L}_{clip} =\frac{e^{cos(\Psi(T_{u}), \Phi(I_{g}))}}{\sum _{i\in\left \{ n, u \right \} } e^{cos(\Psi(T_{i}), \Phi(I_{g}))}}, 
\end{equation}
where $I_{g}$ is the image generated by the diffusion model.

Substituting CLIP-Classifier $\mathcal{L}_{clip}$ into Eq. (\ref{eq:noise_scale}), the new $\epsilon_{\theta}(x_{t},y_{1},y_{2},t)$ is:
\begin{equation}
\label{eq:noise_scale_clip}
\begin{aligned}
    \epsilon_{\theta}(x_{t},y_{1},y_{2},t)&=\epsilon_{\theta}(x_{t},t)-\lambda  \sqrt{1-\bar{\alpha}_{t}}\bigtriangledown\log p(y_{1}|x_{t})\\
&-(1-\lambda  )\sqrt{1-\bar{\alpha}_{t}}\bigtriangledown\log \mathcal{L}_{clip}(y_{2},x_{t}),
\end{aligned}
\end{equation}
where due to the properties of the cosine function, $y_{2}=T_{u}$. Next, we will explore the details of the CLIP-Classifier $\mathcal{L}_{clip}$ guiding the generation process.

\begin{figure}[t]
\centering
	\includegraphics[width=1\linewidth]{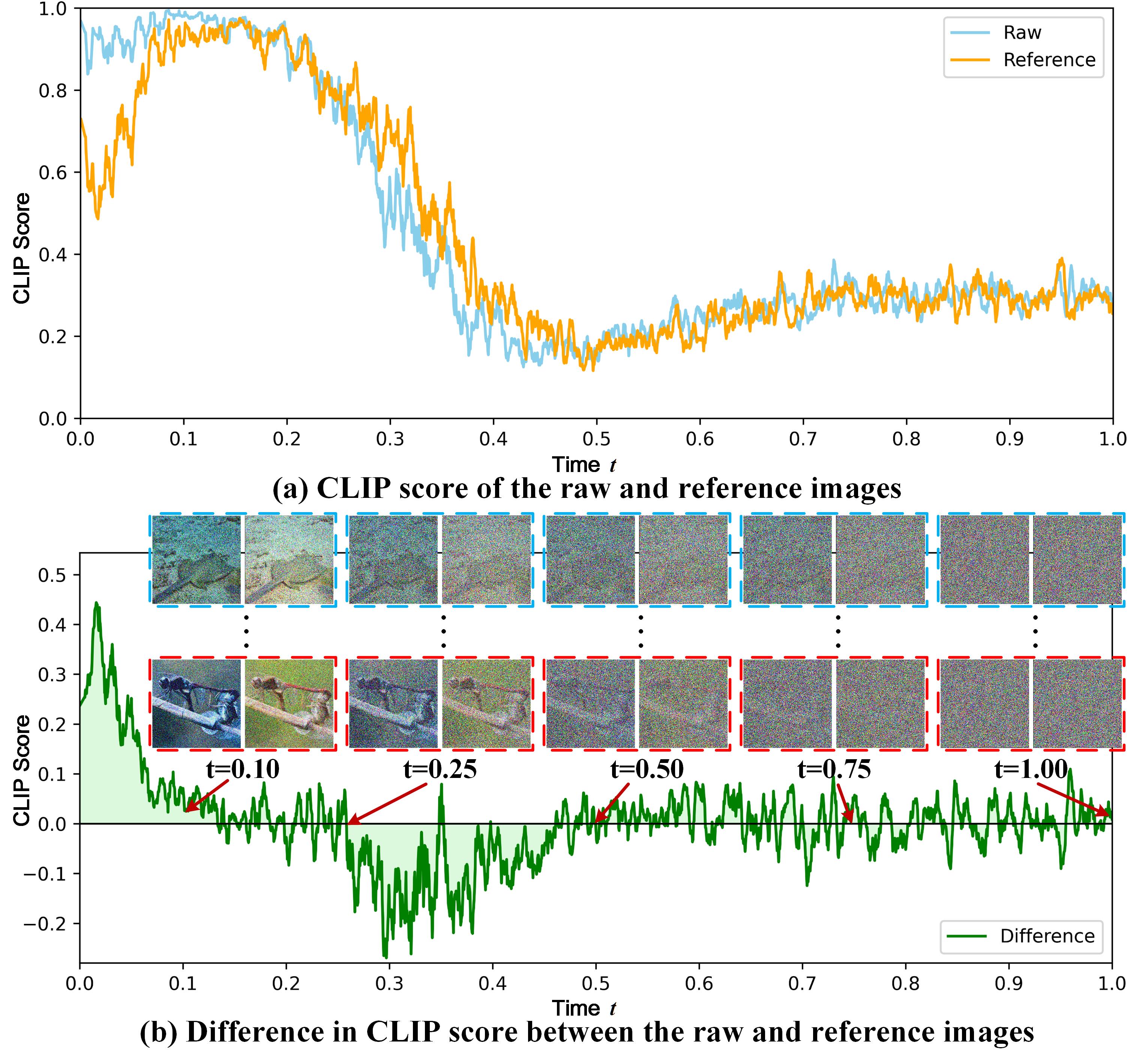}
    \caption{Qualitative analysis of the effectiveness of the CLIP-Classifier. (a) CLIP score of the intermediate variable $x_{t}$ of the image versus time $t$. The time step is set to 2000. The CLIP score curves of the raw and reference images are plotted separately. (b) We calculate the difference in CLIP score between the raw and reference images, then plot the difference curve and label several key time points on the curve.}

     % According to Eq.~\ref{eq:x_t_produce}, we obtain the sequence of intermediate variables $(x_{0},x_{1},\dots,x_{T})$ for each of the raw and reference images. 
    \label{Fig:clip_score_d}
    \vspace{-1em}
\end{figure}

\subsection{New Fine-Tuning Strategy}\label{se:Fine-Tuning Strategy}
The forward process of the diffusion model is actually a process in which image semantic information is continuously obscured by noise, always starting from high-frequency information to low-frequency information. In contrast, the reverse generation process smoothly converts an initial Gaussian noise to a realistic image sample through iterative denoising, working from low-frequency information to high-frequency information. For special image-to-image tasks that require consistent content, such as super-resolution~\cite{saharia2022image}, image restoration~\cite{luo2023image}, repaint~\cite{lugmayr2022repaint}, etc., the input image and the corresponding image share low-frequency semantic information in each transition from $x_{t}$ to $x_{t-1}$ during the generation process~\cite{choi2021ilvr}. Therefore, with the semantic information shared by the input and output images in the low-frequency region, such as the overall structure, image editing and enhancement based on image-to-image diffusion model can start from the high-frequency region instead of the pure Gaussian noise. For instance, SDEdit~\cite{meng2021sdedit} is to “hijack” the generation process, adding a suitable amount of noise to smooth out high-frequency semantic details, while still preserving the overall structure of the input image, and then this noisy input is used to initialize the SDE to obtain a denoised result that is faithful to the user guidance input.\par

Obviously, in our case, the underwater degradation images and the reference images share low-frequency semantic information, so we can introduce new conditions at intermediate moments $x_{t}$ to guide the diffusion process (see in Fig.~\ref{Fig:clip_guidance}). In addition, we further analyze details of how the CLIP-Classifier $\mathcal{L}_{clip}$ guides the generation process. Given a batch of data pairs, we first use Eq. (\ref{eq:x_t_produce}) to produce a sequence of intermediate variables $(x_{0},x_{1},\dots,x_{T})$ for each image by perturbing images with Gaussian noise at time $t\in (0,1)$. Then we employ CLIP-Classifier to calculate the classification probability (CLIP score) of the intermediate variables as well as the score difference. As illustrated in Fig.~\ref{Fig:clip_score_d} (a), the randomness of the perturbation noise directly causes the CLIP score curves to vibrate violently in small intervals, but the overall trend of the curves is still pronounced. From Fig.~\ref{Fig:clip_score_d} (b), we observe that the CLIP-Classifier $\mathcal{L}_{clip}$ mainly works in the discontinuous high-frequency region where $t\in (0,0.10)\cup(0.25,0.50)$, which is consistent with the properties of the image-to-image diffusion model. Based on this finding, we only need to guide the pre-trained diffusion model in the high-frequency region during the fine-tuning process. This new strategy improves the fine-tuning speed compared to the traditional strategy and can increase the speed by up to 10 times, as detailed in Section~\ref{se:ablation}. 

% We find that the CLIP-classifier $\mathcal{L}_{clip}$ mainly acts in the high-frequency region during the fine-tuning process, and We only need to guide the pre-trained diffusion model in the high frequency region. The new strategy makes the fine-tuning process ten times faster than traditional methods.

\section{Experiments}
\subsection{Benchmarks and Evaluation Metrics}
\textbf{Prior Knowledge.} It is well known that real reference images are difficult to obtain for the UIE task. We employ color transfer to degrade the in-air natural images of INaturalist 2021 (mini)~\cite{van2018inaturalist} into the underwater domain to obtain synthetic paired datasets, called \textbf{UIE-air dataset} (see in Fig.~\ref{Fig:pipeline} (a)). INaturalist 2021 (mini) has $10,000$ categories with $50$ images in each category, totalling $500$k images. The image-to-image diffusion model is trained from scratch on this dataset to learn the prior knowledge of mapping transitions from the real underwater degradation domain to the in-air natural domain.\par

\textbf{Dataset.} The SUIM-E~\cite{qi2022sguie} has only $1,635$ images, while UIEB~\cite{li2019underwater} has only $890$ images, which is significantly smaller than the INaturalist dataset used to pre-train the diffusion model. Therefore, we utilize the pre-divided \textbf{SUIM-E-Train} set, which consists of $1,525$ paired underwater images, along with $800$ paired images randomly selected from UIEB, as the training set to fine-tune our model. The test set from SUIM-E and the remaining $90$ images from UIEB are combined to form a test set of $200$ images, called \textbf{T200}.\par

We also conduce tests on the Challenging set of UIEB, which contains $60$ tough cases for enhancement without references, called \textbf{C60}. Additionally, we further verify the generalization and robustness of our model on other commonly used underwater image datasets, which also lack reference images. \textbf{SQUID}~\cite{berman2020underwater} contains $57$ images taken in different locations, enabling a rigorous evaluation of restoration algorithms on underwater images. \textbf{Color-Checker7}~\cite{sharma2005ciede2000} covers $7$ underwater images taken with different cameras in a shallow swimming pool, which is also used to evaluate the accuracy of the proposed CLIP-UIE for color correction.\par

\textbf{Evaluation Metrics.} Since not all datasets have reference images, we divide five commonly used evaluation metrics into two groups for comprehensive quantitative evaluation. For full-reference evaluation metrics, we use Peak Signal to Noise Ratio (\textbf{PSNR}) and Structural Similarity (\textbf{SSIM})~\cite{wang2004image} to measure the consistency between the enhancement and the provided ground-truth images. For non-reference evaluation metrics, we employ three measures: Underwater Image Quality Measure (\textbf{UIQM})~\cite{panetta2015human}, which considers underwater contrast, colorfulness, and sharpness in HSV space; Underwater Color Image Quality Evaluation (\textbf{UCIQE})~\cite{yang2015underwater}, which assesses chroma, contrast and saturation to quantify nonuniform color cast, blurring and low-contrast in degraded underwater images; and Cumulative Probability of Blur Detection (\textbf{CPBD})~\cite{narvekar2011no}, which utilizes a probabilistic model to estimate the probability of detecting blur at each edge in the image.

\begin{table*}[]
 \caption{Quantitative comparison on T200, C60, SQUID and Color-Check7 datasets. Two reference-based metrics (i.e., PSNR and SSIM) and three Non-reference quality metrics (i.e., UIQM, UCIQE and CPBD) are adopted. Double horizontal lines separate the traditional and deep learning-based approaches. The best scores and second-best scores are marked in \textcolor{red}{Red} and \textcolor{blue}{Blue}, respectively. (Best viewed in color) \label{tab:compare_methods}} 
 \fontsize{9pt}{10pt}\selectfont
 \renewcommand{\arraystretch}{1.3}  
 \setlength\tabcolsep{1.3pt}
 \centering
 \scalebox{1}{
\begin{tabular}{c|ccccc|ccc|ccc|ccc}
\hline
\textbf{Datasets}           & \multicolumn{5}{c|}{T200}                                                                                                                                 & \multicolumn{3}{c|}{C60}                                                                   & \multicolumn{3}{c|}{SQUID}                                                                 & \multicolumn{3}{c}{Color-Checker7}                                                           \\ \hline
Methods              & PNSR$\uparrow$                          & SSIM$\uparrow$                         & UIQM$\uparrow$                         & UCIQE$\uparrow$                        & CPBD$\uparrow$                         & UIQM$\uparrow$                         & UCIQE$\uparrow$                        & CPBD$\uparrow$                         & UIQM$\uparrow$                         & UCIQE$\uparrow$                        & CPBD$\uparrow$                         & UIQM$\uparrow$                         & UCIQE$\uparrow$                        & CPBD$\uparrow$                         \\ \hline
UDCP                        & 11.803                        & 0.548                        & {\color[HTML]{FE0000} 1.268} & 0.598                        & 0.636                        & {\color[HTML]{FE0000} 0.952} & 0.552                        & 0.551                        & {\color[HTML]{FE0000} 0.680} & {\color[HTML]{FE0000} 0.575} & 0.627                        & {\color[HTML]{FE0000} 1.642} & {\color[HTML]{FE0000} 0.641} & {\color[HTML]{FE0000} 0.589} \\
ULAP                        & {\color[HTML]{FE0000} 16.570} & {\color[HTML]{FE0000} 0.768} & 0.955                        & {\color[HTML]{FE0000} 0.614} & {\color[HTML]{FE0000} 0.647} & 0.690                        & {\color[HTML]{FE0000} 0.579} & {\color[HTML]{FE0000} 0.560} & 0.466                        & 0.511                        & {\color[HTML]{FE0000} 0.659} & 0.752                        & 0.628                        & 0.557                        \\ \hline\hline
Ucolor                      & 21.907                        & 0.888                        & 0.586                        & 0.587                        & 0.606                        & 0.505                        & 0.553                        & {\color[HTML]{3166FF} 0.512} & 0.182                        & 0.522                        & 0.637                        & 0.693                        & 0.586                        & 0.580                        \\
TCTL-Net                    & 22.403                        & 0.897                        & 0.796                        & 0.608                        & 0.617                        & 0.576                        & {\color[HTML]{3531FF} 0.587} & 0.510                        & 0.133                        & 0.547                        & {\color[HTML]{FE0000} 0.656} & {\color[HTML]{3531FF} 1.092} & 0.613                        & {\color[HTML]{3531FF} 0.583}                        \\
UIEC\textasciicircum{}2-Net & 23.347                        & 0.860                        & {\color[HTML]{3531FF} 0.822} & {\color[HTML]{3531FF} 0.610} & {\color[HTML]{FE0000} 0.631} & {\color[HTML]{3531FF} 0.661} & 0.583                        & {\color[HTML]{FE0000} 0.549} & {\color[HTML]{3531FF} 0.323} & {\color[HTML]{3531FF} 0.571}                        & {\color[HTML]{3531FF} 0.641} & 0.893                        & {\color[HTML]{3531FF} 0.619} & {\color[HTML]{FE0000} 0.617} \\

UDAfomer                    & 25.350                        & 0.921                        & 0.759                        & 0.596                        & 0.606                        & 0.543                        & 0.561                        & 0.502                        & 0.187                        & 0.545                        & 0.607                        & 0.949                        & 0.609                        & 0.570                        \\
{\footnotesize DM\_underwater}              & {\color[HTML]{FE0000} 25.569} & {\color[HTML]{3531FF} 0.931} & 0.797                        & 0.609                        & 0.576                        & 0.647                        & 0.579                        & 0.492                        & 0.242                        & 0.560 & 0.612                        & 0.874                        & 0.614                        & 0.576                        \\

CLIP-UIE                    & {\color[HTML]{3531FF} 25.412} & {\color[HTML]{FE0000} 0.936} & {\color[HTML]{FE0000} 0.981} & {\color[HTML]{FE0000} 0.619} & {\color[HTML]{3531FF} 0.624} & {\color[HTML]{FE0000} 0.754} & {\color[HTML]{FE0000} 0.588} & 0.497                        & {\color[HTML]{FE0000} 0.424} & {\color[HTML]{FE0000} 0.575} & 0.629                        & {\color[HTML]{FE0000} 1.257} & {\color[HTML]{FE0000} 0.645} & 0.544                        \\ \hline
\end{tabular}
 }
 \vspace{-1em}
 \end{table*}

 \begin{figure*}[t]
\centering
	\includegraphics[width=1\linewidth]{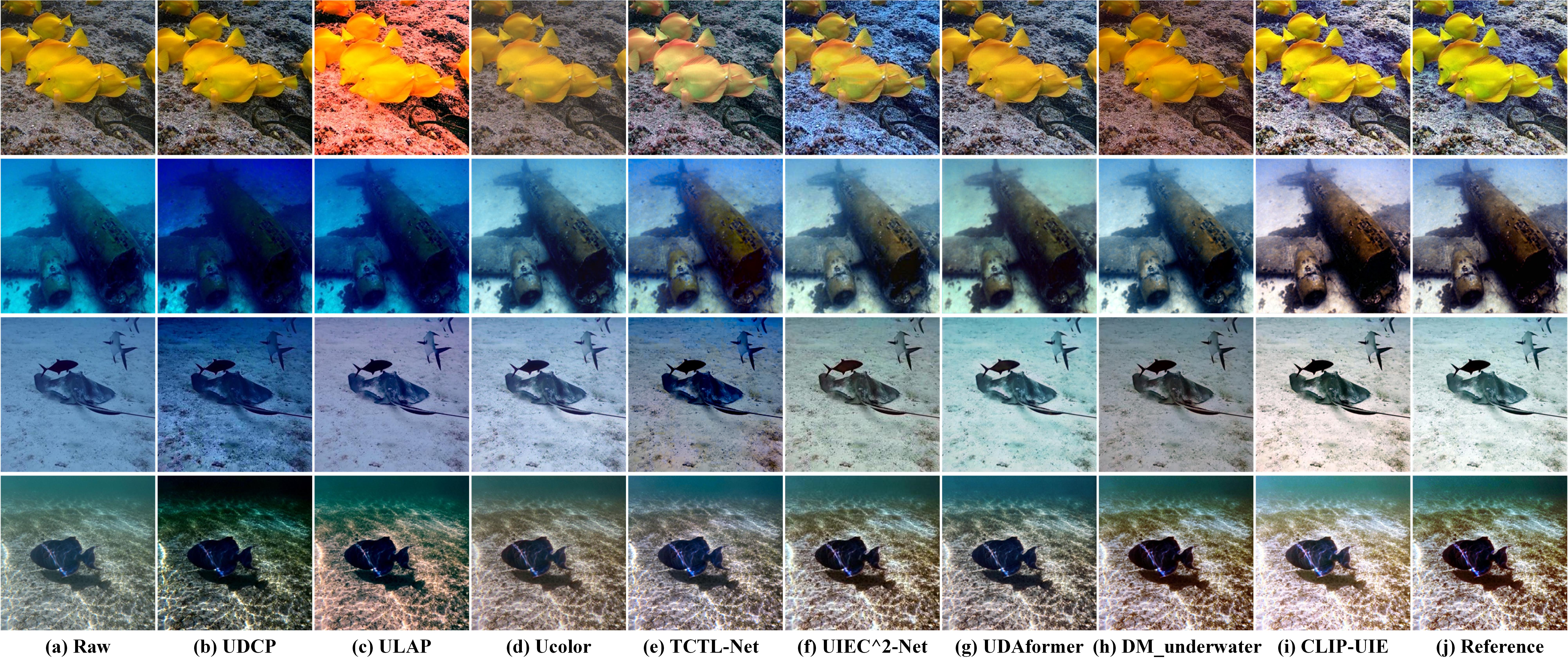}
    \caption{Visual comparisons on underwater images from T200 dataset. From left to right are raw underwater images and the results of UDCP~\cite{drews2013transmission}, ULAP~\cite{song2018rapid}, Ucolor~\cite{li2021underwater}, TCTL-Net~\cite{li2023tctl}, UIEC\^{}2-Net~\cite{wang2021uiec}, UDAformer~\cite{shen2023udaformer}, DM\_underwater~\cite{tang2023underwater}, the proposed CLIP-UIE, and reference images are presented, respectively.}
    \label{Fig:T200}
\end{figure*}

 \begin{figure*}[t]
\centering
	\includegraphics[width=1\linewidth]{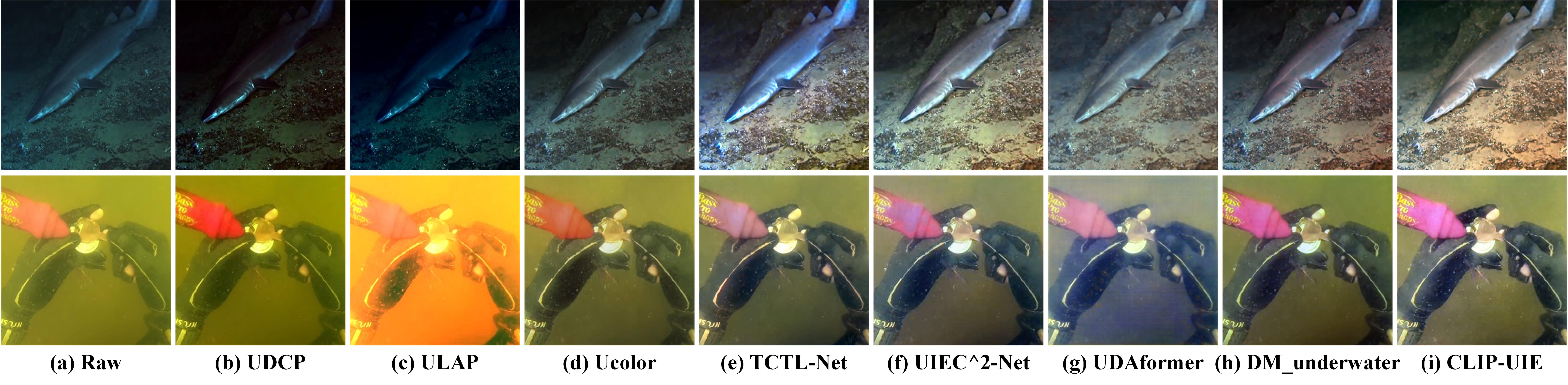}
    \caption{Visual comparisons on challenging underwater images from C60 dataset. From left to right are raw underwater images and the results of UDCP~\cite{drews2013transmission}, ULAP~\cite{song2018rapid}, Ucolor~\cite{li2021underwater}, TCTL-Net~\cite{li2023tctl}, UIEC\^{}2-Net~\cite{wang2021uiec}, UDAformer~\cite{shen2023udaformer}, DM\_underwater~\cite{tang2023underwater} and the proposed CLIP-UIE are presented, respectively.}
    \label{Fig:C60}
\end{figure*}
 \begin{figure*}[t]
\centering
	\includegraphics[width=1\linewidth]{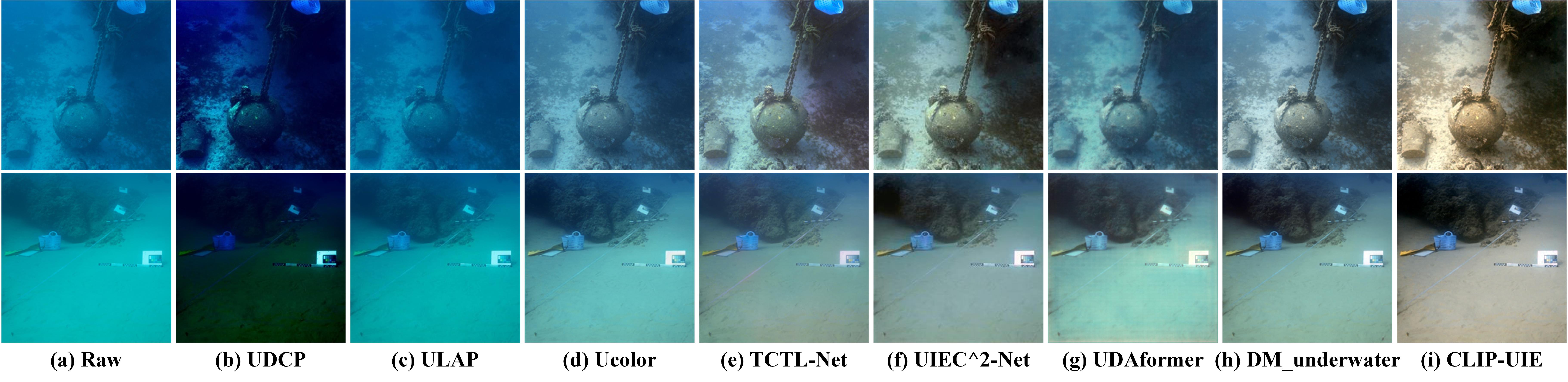}
    \caption{Visual comparisons on challenging underwater images from SQUID dataset. From left to right are raw underwater images and the results of UDCP~\cite{drews2013transmission}, ULAP~\cite{song2018rapid}, Ucolor~\cite{li2021underwater}, TCTL-Net~\cite{li2023tctl}, UIEC\^{}2-Net~\cite{wang2021uiec}, UDAformer~\cite{shen2023udaformer}, DM\_underwater~\cite{tang2023underwater} and the proposed CLIP-UIE are presented, respectively.}
    \label{Fig:SQUID}
\end{figure*}
 \begin{figure*}[t]
\centering
	\includegraphics[width=1\linewidth]{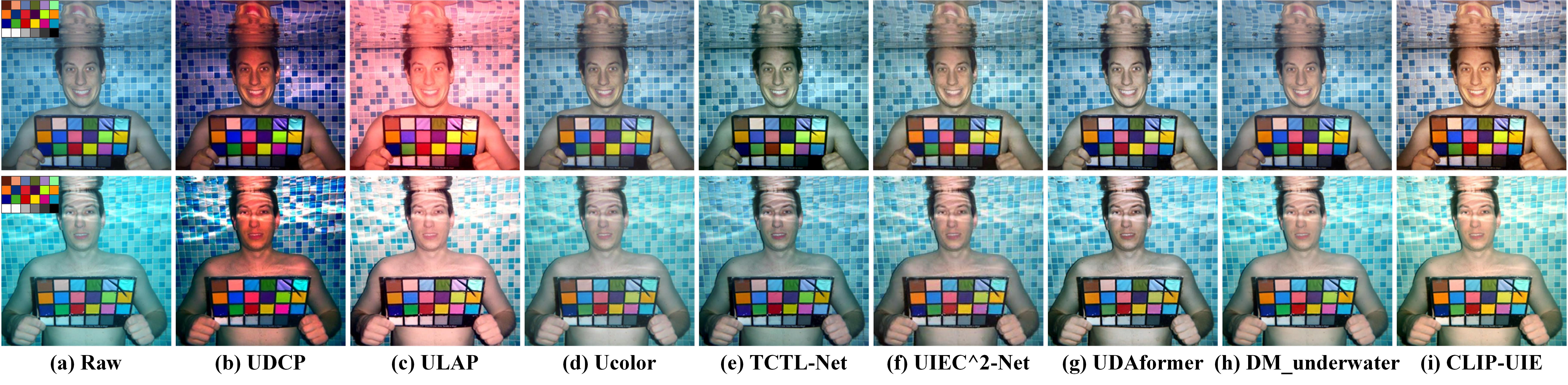}
    \caption{Visual comparisons on underwater images from Color-Checker7 dataset. From left to right are raw underwater images and the results of UDCP~\cite{drews2013transmission}, ULAP~\cite{song2018rapid}, Ucolor~\cite{li2021underwater}, TCTL-Net~\cite{li2023tctl}, UIEC\^{}2-Net~\cite{wang2021uiec}, UDAformer~\cite{shen2023udaformer}, DM\_underwater~\cite{tang2023underwater} and the proposed CLIP-UIE are presented, respectively. The standard color card is appended to the upper left corner of the first column.}
    \label{Fig:Color-Check7}
\end{figure*}

\subsection{Implementation Details}
We implement our method with the PyTorch platform. For training, we initialize all the weights of each layer with Xavier initialization~\cite{glorot2010understanding} and employ the Adam optimizer with a learning rate of $3e$-$6$ with the scheduler of linear decay to zero. Moreover, batch size is set to $8$, and the time step of the diffusion model is set to $2000$ (namely $T=2000$). To ensure a fair comparison, images for all compared methods were resized to $256\times 256$ during both training and testing. Additionally, our data augmentation techniques include random rotation and random horizontal flips. The experiments are conducted on a consistent hardware setup comprising a single NVIDIA RTX 3090 GPU, a $2.3$GHz Intel Xeon processor, $64$GB RAM, and the Ubuntu 18.04 operating system platform. The pre-trained model we used has been trained from scratch for $300$ hours on \textbf{UIE-air dataset}.

\subsection{Compared Methods}
We compare our method CLIP-UIE with seven state-of-the-art UIE approaches, including two typical traditional methods (UDCP~\cite{drews2013transmission} and ULAP~\cite{song2018rapid}), and five deep learning-based methods (Ucolor~\cite{li2021underwater}, TCTL-Net~\cite{li2023tctl}, UIEC\^{}2-Net~\cite{wang2021uiec}, UDAformer~\cite{shen2023udaformer}, DM\_underwater~\cite{tang2023underwater}). The first two UIE methods are the most representative traditional methods that have been widely adopted in recent years. The deep learning-based approaches compared here have achieved state-of-the-art (SOTA) performance over the past years. Among them, UDAformer~\cite{shen2023udaformer} utilizes a dual attention transformer structure to establish the enhancing framework, and DM\_underwater~\cite{tang2023underwater} represents the pioneering UIE method based on diffusion model. They demonstrate the application of the SOTA technologies in the UIE tasks. To ensure a fair and rigorous comparison, we employed the same training set from both the SUIM-E and UIEB datasets to fine-tune or further train all deep learning-based methods. Since the training phase of Ucolor ~\cite{li2021underwater} requires specified image sizes, we only use $256\times 256$ image sizes for the test phase. Furthermore, other comparison methods also adopt $256\times 256$ image size in both training and testing phases. Additionally, all the comparison methods utilize the source codes released by their respective authors to generate their results and strictly adhere to identical experimental settings throughout all evaluations.\par

\subsection{Comparison on Visual Quality of Enhancement}
The quantitative comparisons on the T200, C60, SQUID and Color-Check7 datasets are reported in Table \ref{tab:compare_methods}. We mainly use PSNR and SSIM as our quantitative indices for full-reference T200 datasets, and UIQM, UCIQE and CPBD for non-reference C60, SQUID and Color-Checker7 datasets. The results in Table \ref{tab:compare_methods} show that our CLIP-UIE outperforms the other compared methods, and achieve state-of-the-art performance in terms of image quality evaluation metrics on UIE task, particularly with SSIM, UIQM and UCIQE reaching 0.936, 0.981 and 0.619 respectively. This verifies that our method can produce more natural enhanced images, which better corrects degraded colors and maps underwater images to the in-air natural domain. To better validate the superiority of our methods, Fig.~\ref{Fig:T200} shows the visual results of different comparison methods on T200 dataset. The traditional methods like UDCP and ULAP usually cause color distortions and damage image detail (see the examples in Fig.~\ref{Fig:T200} (b, c)), albeit with high non-reference metrics, which also occurs in other datasets. The deep learning-based methods can produce pleasing visual enhancement due to the direct use of reference images as constraints during the training phrase. However, this limits the enhanced image to a manually selected reference domain, and some methods have very similar enhancement results and do not outperform the reference (e.g. (d - h) in the second and fourth rows of Fig.~\ref{Fig:T200}). On the contrary, our CLIP-UIE introduces the prior knowledge of in-air natural domain through CLIP-Classifier, which can produce more natural images, breaking through the limitation imposed by reference domain to a certain extent, as shown in Fig.~\ref{Fig:T200} (i).\par

To further verify the effectiveness and robustness of our method, we conducted experiments on C60, SQUID and Color-Checker7 datasets. (1) The visual comparisons with other methods on the C60 dataset are shown in the Fig.~\ref{Fig:C60}. The underwater images suffer from lower contrast and more severe color casts (see the examples in the first row of Fig.~\ref{Fig:C60}), which raises great challenges for the enhancement methods. By observation, the deep learning-based methods show better robustness and generalisation than the traditional methods. As a comparison, our method has better performance in color correction and detail restoration. Specially, our method achieves a 0.09 improvement over the second best method, UIEC\^{}2-Net~\cite{wang2021uiec}, in UIQM metrics. (2) On the SQUID dataset, our CLIP-UIE outperforms all other deep learning-based methods and achieves optimal scores in terms of underwater image quality non-reference evaluation metrics including UIQM and UCIQE, with 0.424 and 0.575, respectively. (see in Table~\ref{tab:compare_methods}). As shown in Fig. \ref{Fig:SQUID}, our method performs more natural visual enhancement for underwater images taken at different depths, which verifies the robustness of the proposed CLIP-UIE. This phenomenon can be attributed to the conditional generation is jointly controlled by the source image $y_{1}$ and in-air natural domain $y_{2}$ introduced by CLIP-Classifier (the process is shown in Fig.~\ref{Fig:clip_guidance}), thereby the multi-conditions guidance enhances generation quality. (3) We further conduce comparative tests on the Color-Checker7 dataset to evaluate the generalization and robustness of our model. Fig. \ref{Fig:Color-Check7} visually illustrates the difference between output images of CLIP-UIE and other methods. In the first row of Fig.~\ref{Fig:Color-Check7}, the difference lies in the correction of skin color, where the image generated by CLIP-UIE is closer to the real colors. In the second row of Fig.~\ref{Fig:Color-Check7}, the ULAP~\cite{song2018rapid} produce the enhanced image with bright background lighting, but there is still a color deviation due to the introduction of excessive red components. Comparing to deep learning-based methods, the results produced by our method are brighter and more visually appealing.\par
\begin{table}[]
 \caption{The color dissimilarity comparisons of different methods on color-check7 in terms of the CIEDE2000. The smaller the value, the better the performance. The Top Three Scores are in \textcolor{red}{Red}, \textcolor{blue}{Blue}, \textcolor{green}{Green}. (Best Viewed in Color) \label{tab:color_restoration}} 
 \fontsize{9pt}{10pt}\selectfont
 \renewcommand{\arraystretch}{1.5}  
 \setlength\tabcolsep{2pt}
 \centering
 \scalebox{1}{
\begin{tabular}{c|ccccccc}
\hline
Method                      & D10                          & Z33                          & T6000                        & T8000                        & TS1                          & W60                          & W80                          \\ \hline\hline
Input                       & 12.91                        & 16.65                        & 14.99                        & 19.30                        & 16.15                        & 11.97                        & 14.12                        \\
UDCP                        & 25.76                        & 24.54                        & 21.79                        & 32.99                        & 29.04                        & 23.57                        & 24.54                        \\
ULAP                        & 26.46                        & 17.79                        & 18.25                        & 25.01                        & 19.26                        & 24.75                        & 18.36                        \\
Ucolor                      & {\color[HTML]{3531FF} 11.66} & 15.11                        & 17.24                        & {\color[HTML]{FE0000} 18.59} & {\color[HTML]{32CB00} 15.93} & {\color[HTML]{FE0000} 10.15} & 13.58                        \\
TCTL-Net                    & 12.22                        & {\color[HTML]{009901} 14.96} & {\color[HTML]{3531FF} 13.51} & 20.33                        & 16.34                        & {\color[HTML]{3531FF} 10.70} & 16.64                        \\
UIEC\textasciicircum{}2-Net & 17.81                        & 21.86                        & 17.55                        & {\color[HTML]{3531FF} 19.61} & 20.82                        & 16.11                        & {\color[HTML]{3531FF} 11.59} \\

UDAformer                   & {\color[HTML]{009901} 11.99} & 16.15                        & {\color[HTML]{009901} 13.98} & 21.62                        & {\color[HTML]{FE0000} 14.11} & {\color[HTML]{000000} 12.09} & {\color[HTML]{FE0000} 11.10} \\
{\footnotesize DM\_underwater}              & 13.19                        & {\color[HTML]{FE0000} 13.97} & 16.78                        & {\color[HTML]{009901} 19.68} & 17.89                        & 13.64                        & {\color[HTML]{009901} 12.09} \\
CLIP-UIE                    & {\color[HTML]{FE0000} 10.18} & {\color[HTML]{3531FF} 14.43} & {\color[HTML]{FE0000} 12.10} & 20.36                        & {\color[HTML]{3531FF} 15.52} & {\color[HTML]{009901} 11.56} & 12.39                        \\ \hline
\end{tabular}
 }
 \vspace{-1em}
 \end{table}

\begin{table}[]
 \caption{Quantitative results of ablation study on T200 datasets. The best scores are in \textcolor{red}{Red}. (Best viewed in color)  \label{tab:Ablation_only2}} 
 \fontsize{9pt}{10pt}\selectfont
 \renewcommand{\arraystretch}{1.5}  
 \setlength\tabcolsep{2pt}
 \centering
 \scalebox{1}{
\begin{tabular}{c|ccccc}
\hline
Models & PNSR$\uparrow$   & SSIM$\uparrow$  & UIQM$\uparrow$  & UCIQE$\uparrow$ &CPBD$\uparrow$  \\ \hline\hline
{\color[HTML]{000000} -w/o-CLIP}      & {\color[HTML]{FE0000} 26.174} & {\color[HTML]{FE0000} 0.943} & {\color[HTML]{000000} 0.783} & {\color[HTML]{000000} 0.585} & {\color[HTML]{000000} 0.607} \\
{\color[HTML]{000000} CLIP-UIE}       & {\color[HTML]{000000} 25.412} & {\color[HTML]{000000} 0.936} & {\color[HTML]{FE0000} 0.981} & {\color[HTML]{FE0000} 0.619} & {\color[HTML]{FE0000} 0.624} \\ \hline
\end{tabular}
 }
 \vspace{-1em}
 \end{table}

 \begin{figure}[t]
\centering
	\includegraphics[width=0.9\linewidth]{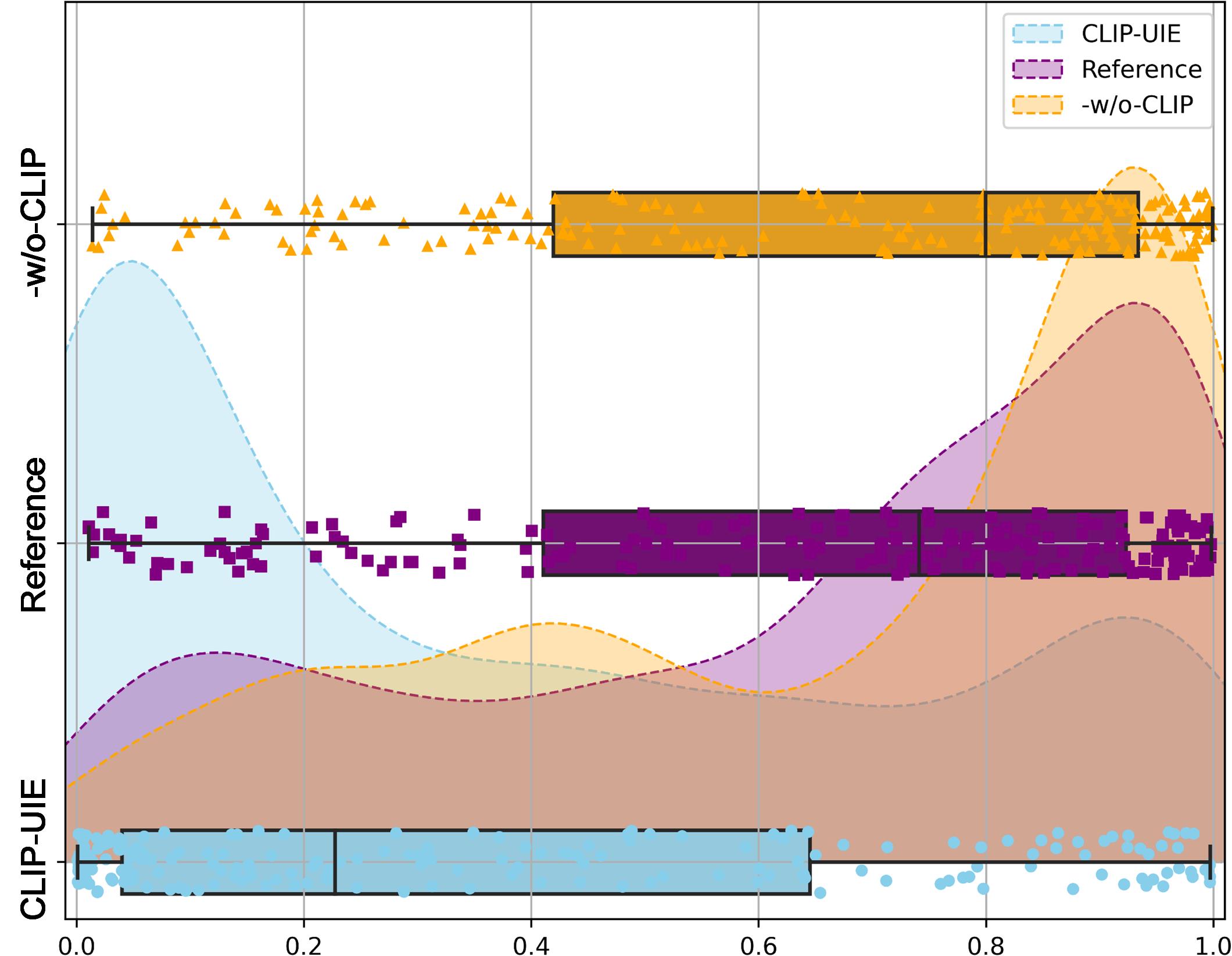}
    \caption{The distribution of similarity score (CLIP score) between the learned in-air natural image prompt and images across the T200 dataset. The point-box plot and kernel density estimation curve of the reference images and the enhancements with the CLIP-UIE-w/o-CLIP and complete CLIP-UIE are presented. The closer the point-box plot is to the left, the closer it is to the in-air natural domain. }
    \label{Fig:clip_score}
\end{figure}

 \begin{figure}[t]
\centering
	\includegraphics[width=1\linewidth]{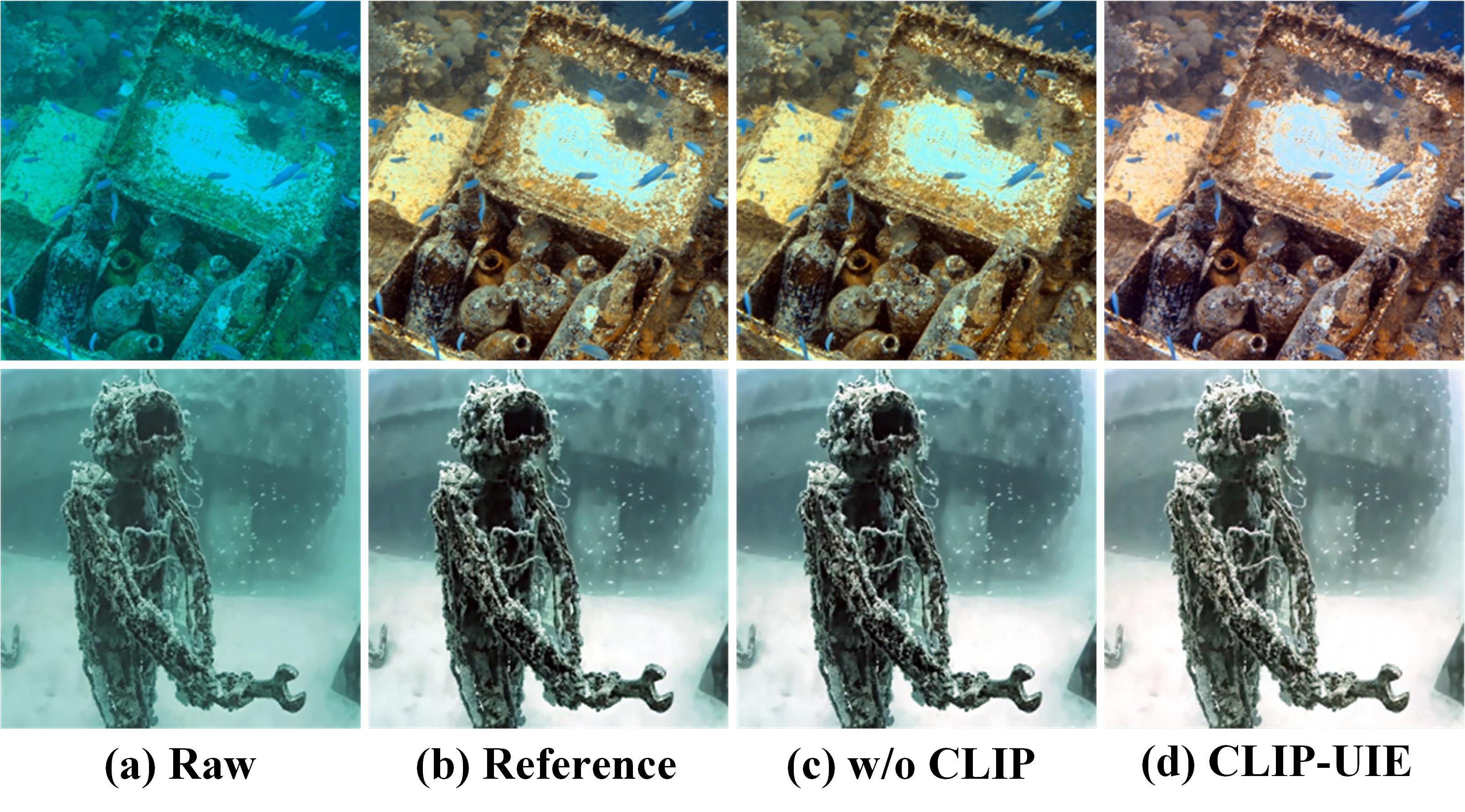}
    \caption{Ablation study of the effectiveness of CLIP-Classifier. From left to right are raw underwater images, reference images, the enhancements with the CLIP-UIE-w/o-CLIP and complete CLIP-UIE are presented, respectively. With CLIP-Classifier, CLIP-UIE breaks through the limitations of the reference domain and achieves more natural enhancement results.}
    \label{Fig:ablation_T200}
\end{figure}

\subsection{Comparison on Color Correction Performance}
In order to further illustrate the performance of the proposed CLIP-UIE for color restoration of underwater images, we conduced detailed color dissimilarity comparisons on the Color-Checker7~\cite{sharma2005ciede2000} dataset, which is photographed in a swimming pool with divers holding standard color cards. Following Ancuti et al.~\cite{song2018rapid}, we use CIEDE2000~\cite{sharma2005ciede2000} to measure the relative differences between the corresponding color patches of the Macbeth color checker~\cite{pascale2006rgb} and the enhancement results of these comparison state-of-the-art methods. For better illustration, a standard color card is appended to the upper left corner of the first column of Fig.~\ref{Fig:Color-Check7}. The results, as presented in Table~\ref{tab:color_restoration}, indicate that our CLIP-UIE can improve the performance in recovering ground-truth colors, but not very noticeably. However, the color performance of the image T8000 is even worse than the input. We believe that this is due to the image encoder $\Phi(\cdot)$ compresses images into high-dimensional features, and CLIP-Classifier pays more attention to global features and ignores the color restoration of local color blocks during generation process. As shown in the first row of Fig.~\ref{Fig:Color-Check7}, the result of our method presents better brightness and visual perception, but does not achieve a significant improvement in color correction.

\subsection{Ablation Study}\label{se:ablation}
\textbf{Effectiveness of CLIP-Classifier.} To Investigate the impact of the CLIP-Classifier on our method, we fine-tune our conditional diffusion model on the same training data without CLIP-Classifier (\textbf{w/o-CLIP}). The scores of the main evaluation metrics on T200 datasets are given in Table  \ref{tab:Ablation_only2}. The results indicate that CLIP-Classifier guides the diffusion model toward the in-air natural domain, leading to a decrease in the full-reference metrics and an increase in the non-reference metrics. Besides, we use CLIP-Classifier to evaluate the similarity (CLIP score) between the in-air natural domain and enhanced results, and the corresponding point-box plot and kernel density estimation (KDE) curve are shown in Fig.~\ref{Fig:clip_score}.  The distributions of the \textbf{w/o-CLIP} and the \textbf{Reference} are very similar (see KDE in Fig.~\ref{Fig:clip_score}), which indicates that its enhancement performance is constrained by the reference domain. On the contrary, due to the introduction of the prior knowledge of in-air natural domain through CLIP-Classifier, our CLIP-UIE breaks through this limitation imposed by reference domain to a certain extent, keeping the distribution of the \textbf{CLIP-UIE} far away from the reference domain and close to the in-air natural domain, reducing subjective preferences.\par

Visual comparison illustrated in Fig.~\ref{Fig:ablation_T200} can also significantly support the above points. Compared with its ablated versions, the proposed CLIP-UIE achieves more natural enhancement results with the guidance of CLIP-Classifier that inherits the prior knowledge of the in-air natural domain. The combination of the reference domain and the in-air natural domain drives the enhancement results closer to real-world natural images (see Fig.~\ref{Fig:clip_guidance} and Fig.~\ref{Fig:ablation_T200} (c, d)).

 \begin{figure*}[t]
\centering
	\includegraphics[width=1\linewidth]{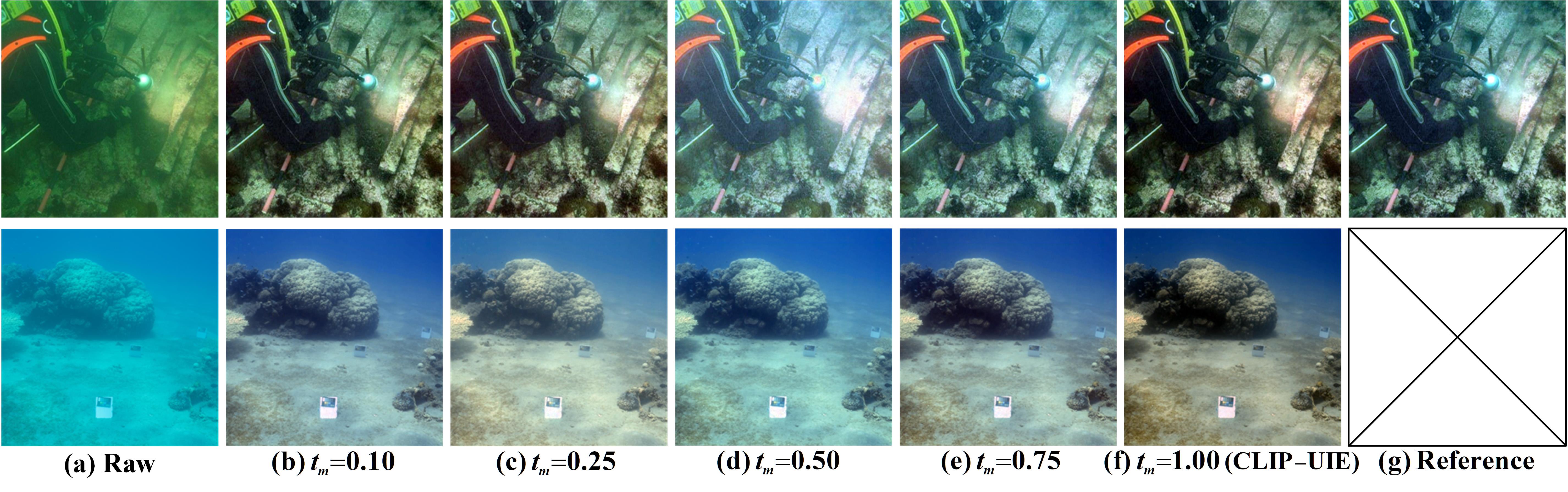}
    \caption{The visual comparisons of the enhancements with different ablated models and the full CLIP-UIE model on the T200 and SQUID dataset. From the left to right, (a) the raw images, the enhancements with ablated models (b) the model $t_{m}$=0.10, (c) the model $t_{m}$=0.25, (d) the model $t_{m}$=0.50, (e) the model $t_{m}$=0.75, (f) the model $t_{m}$=1.00 (the full CLIP-UIE model) and (g) the reference images are presented, respectively. In the second line, the reference image of the SQUID dataset is not available.}
    \label{Fig:ablation_T200_SQUID}
    % The visual comparison of the enhancements with different ablated models and the full CLIP-UIE model (the model $t_{m}$=1.00).
\end{figure*}

\begin{table}[]
 \caption{Efficiency analysis and comparison of new fine-tuning strategy on T200 and SQUID datasets. The best and second-best scores are in \textcolor{red}{Red} and \textcolor{blue}{blue}, respectively. (Best viewed in color)} 
 \fontsize{9pt}{10pt}\selectfont
 \label{tab:Ablation_t}
 \renewcommand{\arraystretch}{1.3}  
 \setlength\tabcolsep{1.3pt}
 \centering
 \scalebox{0.87}{
\begin{tabular}{c|ccccc|ccc}
\hline
Dataset                       & \multicolumn{5}{c|}{T200}                                                                                                                                 & \multicolumn{3}{c}{SQUID}                                                                  \\ \hline
{\color[HTML]{000000} Models} & {\color[HTML]{000000} PNSR}$\uparrow$   & {\color[HTML]{000000} SSIM}$\uparrow$  & {\color[HTML]{000000} UIQM}$\uparrow$  & {\color[HTML]{000000} UCIQE}$\uparrow$ & {\color[HTML]{000000} CPBD}$\uparrow$  & UIQM$\uparrow$                         & UICQE$\uparrow$                        & CPBD$\uparrow$                         \\ \hline\hline
$t_{m}$=0.10 & {\color[HTML]{000000} 24.339} & {\color[HTML]{000000} 0.930} & {\color[HTML]{3531FF} 0.943} & {\color[HTML]{000000} 0.611} & {\color[HTML]{000000} 0.619} & {\color[HTML]{3531FF} 0.417} & {\color[HTML]{333333} 0.550} & {\color[HTML]{FE0000} 0.637} \\
$t_{m}$=0.25 & {\color[HTML]{000000} 23.887} & {\color[HTML]{000000} 0.923} & {\color[HTML]{000000} 0.894} & {\color[HTML]{000000} 0.602} & {\color[HTML]{000000} 0.618} & 0.291                        & 0.541                        & 0.622                        \\
$t_{m}$=0.50 & {\color[HTML]{000000} 24.470} & {\color[HTML]{000000} 0.912} & {\color[HTML]{333333} 0.936} & {\color[HTML]{FE0000} 0.622} & {\color[HTML]{FE0000} 0.651} & 0.376                        & {\color[HTML]{FE0000} 0.583} & {\color[HTML]{3531FF} 0.631} \\
$t_{m}$=0.75 & {\color[HTML]{3531FF} 25.354} & {\color[HTML]{FE0000} 0.946} & 0.924 & {\color[HTML]{000000} 0.608} & {\color[HTML]{3531FF} 0.625} & 0.389                        & 0.559                        & 0.628                        \\
 $t_{m}$=1.00 & {\color[HTML]{FE0000} 25.412} & {\color[HTML]{3531FF} 0.936} & {\color[HTML]{FE0000} 0.981} & {\color[HTML]{3531FF} 0.619} & {\color[HTML]{000000} 0.624} & {\color[HTML]{FE0000} 0.424} & {\color[HTML]{3531FF} 0.575} & 0.629                        \\ \hline
\end{tabular}}
 
 \vspace{-1em}
 \end{table}

\textbf{Fine-Tuning Strategy.} In addition to the observation provided in Section~\ref{se:Fine-Tuning Strategy}, to further validate the new fine-tuning strategy, we provide the quantitative comparison in Table \ref{tab:Ablation_t}. The model $t_{m}$ represents the fact that we only fine-tune the pretrained diffusion model in the range $t\in (0,t_{m} )$ (the training time step $n\in (0,2000\times t_{m})$) (refer to Fig.~\ref{Fig:clip_guidance}). As showm in Table \ref{tab:Ablation_t}, the fine-tuning range of the model $t_{m}$=0.25 is larger than that of the model $t_{m}$=0.10, but the performance does not exceed the latter, especially the main evaluation metrics PSNR and UIQM decreases from 24.339 and 0.943 to 23.887 and 0.894 respectively on the dataset \textbf{T200}. The drop in UIQM on the dataset \textbf{SQUID} is even more dramatic, from 0.417 to 0.291. We believe this may be due to the fact that the image-to-image diffusion model is more sensitive to high-frequency information. The CLIP-Classifier not only does not steer the generation process in the high-frequency range $t\in (0.1,0.25)$, but also interferes with the condition $y_{1}$ (refer to Fig.~\ref{Fig:clip_guidance}). Further comparison of the models $t_{m}$=0.50, $t_{m}$=0.75 and $t_{m}$=1.00 shows that the performance of the models is stable and there is no drastic change in the evaluation metrics. This indicates that the CLIP-Classifier cannot distinguish the image pairs perturbed by Gaussian noise in the low-frequency region $t\in (0.50,1.00)$ and has less impact on the overall generative process. This experimental result aligns with the Fig.~\ref{Fig:clip_score_d} (b). Fig. \ref{Fig:ablation_T200_SQUID} presents the visual comparison between the proposed CLIP-UIE (the model $t_{m}$=1.00) and the other ablated versions. As shown in Fig.~\ref{Fig:ablation_T200_SQUID} (b), for the image-to-image special task, which shares low-frequency semantic information, we are able to adapt to the downstream task by fine-tuning the model only in the range of $t\in (0,0.10)$ and the enhancement results are competitive, which is 10 times faster over the traditional strategy.\par

\section{Discussion and Future works}
As illustrated in Fig.~\ref{Fig:clip_score}, the score distribution of the results produced by the proposed CLIP-UIE is not clustered enough, and some enhanced images are still in the reference domain. Additionally, the score difference curve of the CLIP-Classifier is different from the normal subjective assumption---the more noise, the lower the curve---and the effective range of the classifier is discontinuous $t\in (0,0.10)\cup(0.25,0.50)$, and the image pairs perturbed by Gaussian noise cannot be distinguished when $t\in (0.10,0.25)\cup(0.50,1.00)$. This makes it difficult to precisely control the generation process of diffusion models. In future research, it would be interesting to train an effective range-stable CLIP-Classifier to achieve precise control of the multi-guidance diffusion model, and to make the enhancement results more natural.\par

As a new fine-tuning strategy, the speed is faster than the traditional fine-tuning strategy, but the overall performance is not superior to traditional methods (refer to Table \ref{tab:Ablation_t}). The trade-off between speed and performance is a common challenge for all deep learning-based methods. Exploring how to strike a balance between speed and performance in the fine-tuning of image-to-image diffusion models is also a promising topic. Unfortunately, this fine-tuning strategy is limited to the image-to-image diffusion model, which shares low-frequency semantic information.
\section{conclusion}
In this paper, we have presented a novel framework called CLIP-UIE for underwater image enhancement, which combines the pre-trained diffusion model and CLIP-Classifier to conduct conditional generation. In our framework, the pre-trained diffusion model has learned the prior knowledge of mapping transitions from the underwater degradation domain to the real in-air natural domain. The customized CLIP-Classifier inherits the prior knowledge of the in-air natural domain, which is used to counteract catastrophic forgetting and mode collapse of the diffusion model during fine-tuning, allowing the enhancement results to break through the limitations of the reference domain to a certain extent. Simultaneously, the image-to-image enhancement task requires consistent content between the input and the output images, so that the diffusion model shares low-frequency semantic information, which means that the range of fine-tuning is concentrated in the high-frequency region. Further analysis of CLIP-Classifier shows that it also mainly acts in the high-frequency region, similar to the image-to-image diffusion model. Therefore, we propose a new fine-tuning strategy that guides the pre-trained diffusion model only in the high-frequency region, e.g., the model $t_{m}$=0.10---we only fine-tune in the range $t\in(0,0.10)$---achieves competitive enhancement results, while the speed is 10 times faster than the traditional strategy. Extensive experiments show that CLIP-UIE not only generates enhanced images of superior quality beyond the reference images, but also surpasses other SOTA UIE methods. In the future, we aim to introduce more prior knowledge by classifiers, and further enhance visual quality.
\ifCLASSOPTIONcaptionsoff
  \newpage
\fi

\bibliographystyle{IEEEtran}

\bibliography{reference}


\end{document}